\documentclass[10pt,journal,compsoc]{IEEEtran}
\ifCLASSOPTIONcompsoc
  \usepackage[nocompress]{cite}
\else
  \usepackage{cite}
\fi
\ifCLASSINFOpdf
  \usepackage[pdftex]{graphicx}
\else
\fi
\pdfoutput=1

\usepackage{xcolor}         
\usepackage{colortbl}
\usepackage{booktabs}
\usepackage{graphicx}
\usepackage{subfigure}
\usepackage{amsmath,amssymb,amsfonts}
\usepackage{amssymb}
\usepackage{algorithm} 
\usepackage{algorithmic} 
\usepackage{bm}
\usepackage{ragged2e}
\newcommand{\ie}{{\emph{i.e.}}, }
\newcommand{\eg}{{\emph{e.g.}}, }

\makeatletter
\newcommand{\rmnum}[1]{\romannumeral #1}
\newcommand{\Rmnum}[1]{\expandafter\@slowromancap\romannumeral #1@}
\makeatother

\begin{document}
%
\title{Learngene: Inheriting Condensed Knowledge from the Ancestry Model to Descendant Models}
%
%
%
%

\author{Qiufeng~Wang\IEEEauthorrefmark{1},
	Xu~Yang\IEEEauthorrefmark{1},
	Shuxia~Lin,
	Jing Wang,
	and~Xin~Geng\IEEEauthorrefmark{2},~\IEEEmembership{Senior~Member,~IEEE}
\IEEEcompsocitemizethanks{\IEEEcompsocthanksitem Q.~Wang, X.~Yang, S.~Lin, J.~Wang, and~X.~Geng are with the School of Computer Science and Engineering, and
	Key Laboratory of Computer Network and Information Integration (Southeast University), Ministry of Education, Southeast University, Nanjing
	211189, China. E-mail: \{qfwang, xuyang\_palm, shuxialin, wangjing91, xgeng\}@seu.edu.cn.  \IEEEauthorrefmark{1}. Equal contribution. \IEEEauthorrefmark{2}. Corresponding author. \protect \\
}
\thanks{Manuscript received xxxxxxxx; revised xxxxxxxx.}}

\IEEEtitleabstractindextext{%
\begin{abstract}
\justifying
During the continuous evolution of one organism's ancestry,  its genes accumulate extensive experiences and knowledge, enabling newborn descendants to rapidly adapt to their specific environments. Motivated by this observation, we propose a novel machine learning paradigm \textit{Learngene} to enable learning models to incorporate three key characteristics of genes. (\rmnum{1}) Accumulating: the knowledge is accumulated during the continuous learning of an \textbf{ancestry model}. (\rmnum{2}) Condensing: the extensive accumulated knowledge is condensed into a much more compact information piece, \ie learngene. (\rmnum{3}): Inheriting: the condensed learngene is inherited to make it easier for \textbf{descendant models} to adapt to new environments. Since accumulating has been studied in well-established paradigms like large-scale pre-training and lifelong learning, we focus on condensing and inheriting, which induces three key issues and we provide the preliminary solutions to these issues in this paper: (\rmnum{1}) \textit{Learngene} Form: the learngene is set to a few integral layers that can preserve significance. (\rmnum{2}) \textit{Learngene} Condensing: we identify which layers among the ancestry model have the most similarity as one pseudo descendant model. (\rmnum{3}) \textit{Learngene} Inheriting: to construct distinct descendant models for the specific downstream tasks, we stack some randomly initialized layers to the learngene layers. Extensive experiments across various settings, including using different network architectures like Vision Transformer (ViT) and Convolutional Neural Networks (CNNs) on different datasets, are carried out to confirm four advantages of \textit{Learngene}: it makes the descendant models 1) converge more quickly, 2) exhibit less sensitivity to hyperparameters, 3) perform better, and 4) require fewer training samples to converge. 
\end{abstract}

}

     \maketitle

\IEEEdisplaynontitleabstractindextext

%
\IEEEpeerreviewmaketitle

\IEEEraisesectionheading{\section{Introduction}\label{sec:introduction}}

%
%
%
%

\begin{figure*}[t]
\setlength{\abovecaptionskip}{0pt}
\setlength{\belowcaptionskip}{10pt}
\begin{center}
\centerline{\includegraphics[width=0.9\textwidth,trim=0 0 0 0,clip]{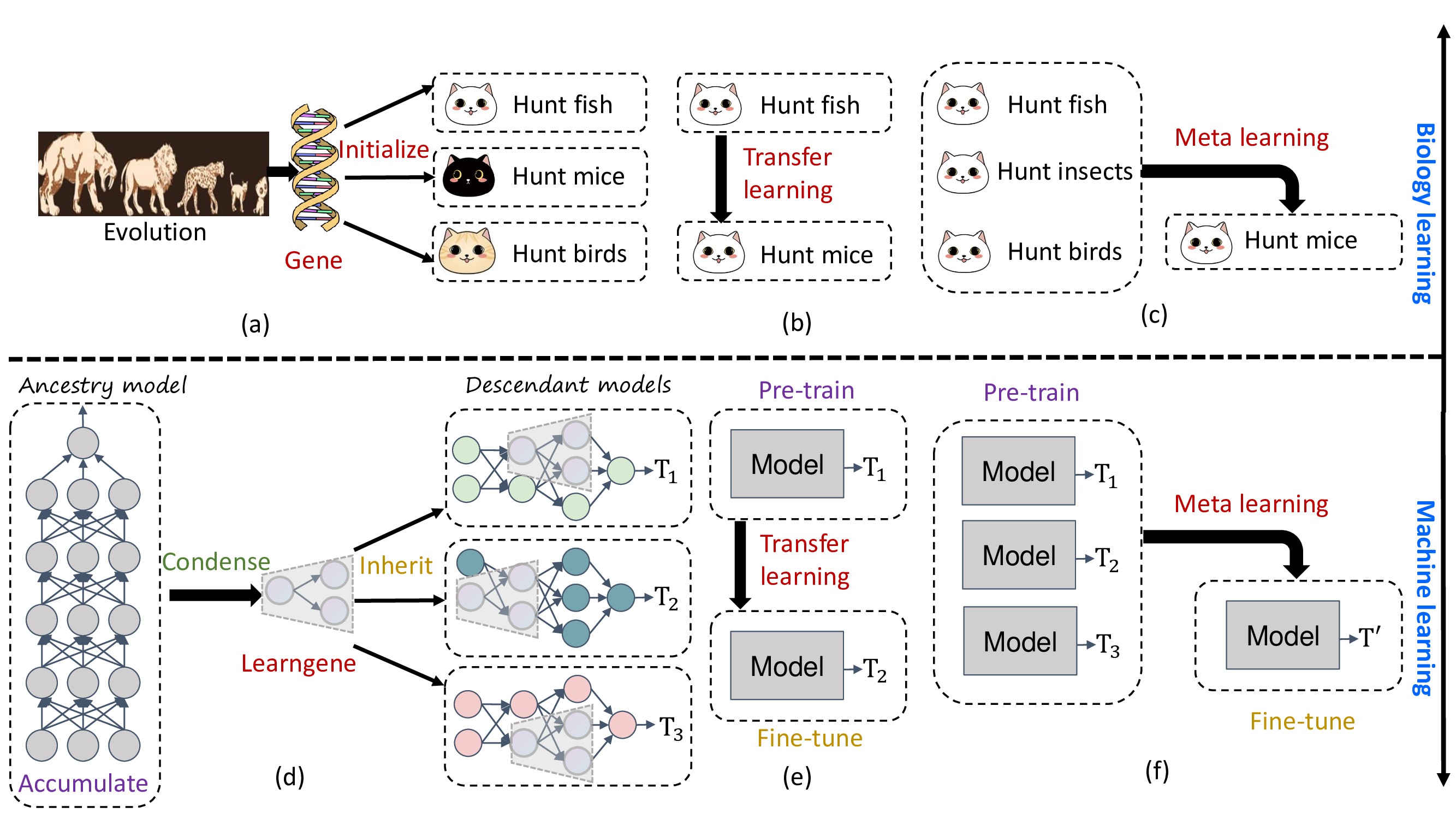}}
\caption{There are several ways of learning in biology,  \eg biological descendants initialized by a gene (a), transfer learning (b), and meta-learning (c).
(d) The learngene is extracted from the ancestry model. Similar to the properties of genes in biology learning, the learngene that represents significant knowledge from the ancestry model is inherited to the lightweight descendant models and enables them to quickly adapt to the target tasks with low-data regimes.  (e) Generally, transfer learning involves adapting the source model to the target task. (f) Typically, meta-learning algorithms learn how to learn across multiple related tasks. }
\label{fig:learngene&relatedwork} 
\end{center}
\end{figure*}

\IEEEPARstart{T}{he} gene of one organism condenses extensive experiences and knowledge, which is accumulated during the continuous evolution~\cite{stearns2000evolution, zador2019critique, hasson2020direct} of this organism's ancestry, into compact information pieces that enable different newborn descendants to quickly adapt to their specific environments. For example, as shown in Figure~\ref{fig:learngene&relatedwork} (a), the hunting skill accumulated by the cat ancestry is condensed into the gene. Then for three descendent cats that live with different prey, the hunting gene effectively initializes their brains to help them quickly learn to hunt these prey through only a few trial and error attempts.

Motivated by this observation, we propose a novel machine learning paradigm named \textit{Learngene} to empower learning models with three key characteristics of the gene. (\rmnum{1}) Accumulating: knowledge is accumulated during the continuous learning of an \textbf{ancestry model}. (\rmnum{2}) Condensing: the extensive accumulated knowledge is condensed into a much more compact piece of information, which is termed learngene.\footnote{To avoid confusion, in the whole paper, we use ``\textit{Learngene}'' to name the proposed learning framework, and use ``learngene'' to term the condensed critical part.} (\rmnum{3}) Inheriting: the condensed learngene is passed on to \textbf{descendant models} to aid quick adaptation to new environments. Such an analogy is sketched in Fig~\ref{fig:learngene&relatedwork} (a) and (d).


Indeed, some approaches like Transfer Learning (TL)
~\cite{zhuang2020comprehensive} or Meta-Learning (ML)~\cite{9428530} have certain overlapping characteristics with \textit{Learngene}, \eg they can also initialize the models for quick adaptation. Conceptually, \textit{Learngene} is distinct, as illustrated by the intuitive comparisons from a biological perspective in Figure~\ref{fig:learngene&relatedwork} (a-c). As demonstrated in (b) and (c), \textbf{the same white cat} transfers/abstracts its hunting experience by TL/ML strategies to quickly learn how to hunt a novel prey like the mice. In contrast, the gene condenses the most stable part during evolution into a much more compact information piece compared with the original organism. Although such a piece is not an integral organism capable of hunting any prey, it is also not biased to one specific organism's hunting experience, which may be trivial or even detrimental for the organisms living in different environments. Thus, the gene, being more stable and demonstrating a stronger generalization ability, aids in initializing \textbf{various cats} to hunt diverse prey without bias towards the experiences of a few cats.

Shifting our viewpoint from biology to machine learning, \textit{Learngene} has analogically different characteristics compared with TL and ML, as demonstrated in Figure~\ref{fig:learngene&relatedwork} (d-f). For TL or ML, they need to reuse the same whole model as the initialization to solve the target tasks. Moreover, the target tasks should resemble the source tasks, otherwise, it requires extensive training samples to adapt to the target tasks. Differently, learngene is a much more compact part that does not have an integral function, while such characteristic endows it with a stronger generalization ability. Specifically, the same learngene can be inherited into the descendant models with different architectures, even the heterogeneous and smaller ones with significantly fewer parameters, enabling rapid adaptation to the corresponding target tasks with just a few training samples. This can occur even when the tasks are dissimilar to the source task, \eg when the data distribution varies between the source and target domains.

It should be stressed that the condensing-inheriting process of \textit{Learngene} is not a trivial variant of the popular pre-training and fine-tuning operation, but holds specific practical values, especially in the era of large-scale pre-trained models. Nowadays, many state-of-the-art (SOTA) models developed by well-funded corporations contain billions of parameters~\cite{dosovitskiy2021an, ramesh2021zero, riquelme2021scaling}. Such models may not be feasible for fine-tuning by mid-sized or small companies to solve specific tasks, let alone by academic research groups with limited computational resources. Thus, it is urgent to have a method that can help these small companies to exploit the SOTA models for solving the specific tasks in a more efficient way, \eg without reusing the entire original model. 

This constitutes another essential motivation for \textit{Learngene}; we aim to extract a compact learngene from a large ancestry model and inherit only this compact part to the descendant models for client companies.
Similarly, learnware~\cite{zhou2016learnware} can provide clients with a machine learning model to solve a target task without training from scratch. However, the models offered by learnware are highly specialized and may not be adaptable to all downstream tasks. More importantly, \textit{Learngene} is designed not to have an integral function like the original model, much like how a gene cannot stand alone as an organism. Such a characteristic can protect the intellectual property of the owner of the ancestry model since the clients are unable to infer the sensitive technical details like the model architecture or the training data~\cite{chakraborty2018adversarial, balle2022reconstructing}. 

As introduced before, \textit{Learngene} involves three key procedures: accumulating, condensing, and inheriting. Among them, numerous studies have proposed effective solutions for the accumulating process, \eg lifelong learning~\cite{Parisi2018ContinualLL}, large-scale pre-training~\cite{Brock2021HighPerformanceLI}. However, to achieve condensing and inheriting, at least three key issues must be addressed: (\rmnum{1}) \textbf{Learngene Form}: What form should \textit{Learngene} take, for example, can it consist of specific integral layers or particular neuron connections? (\rmnum{2}) \textbf{Learngene Condensing}: How can the large model be condensed into the learngene to retain maximum significance without having integral functions for solving any task? (\rmnum{3}) \textbf{Learngene Inheriting}: How can the learngene be used to initialize the descendant models to solve specific tasks?

\begin{figure*}[htb]
\setlength{\abovecaptionskip}{0pt}
\setlength{\belowcaptionskip}{10pt}
\begin{center}
\centerline{\includegraphics[width=0.85\textwidth]{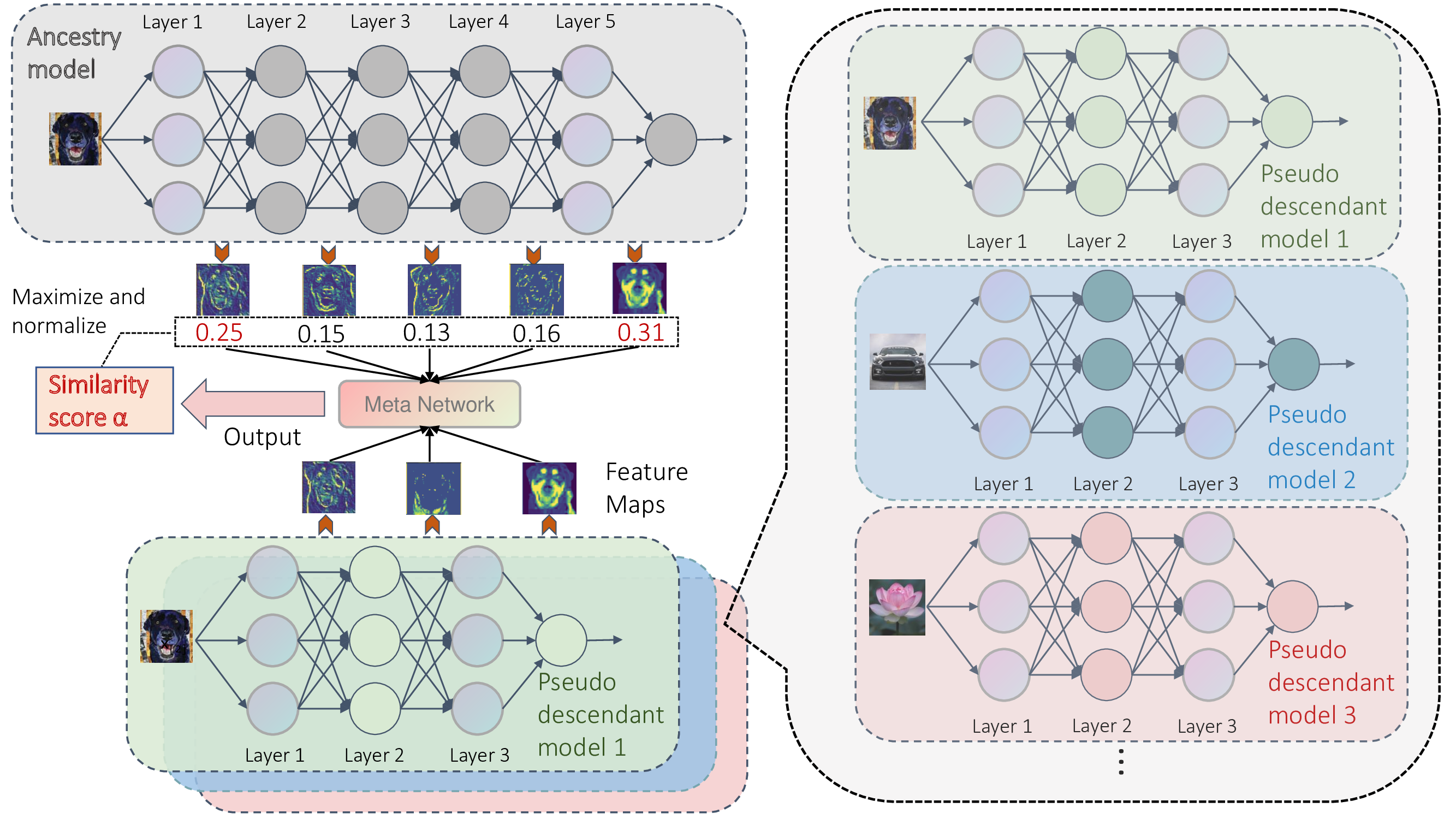}}
\caption{ Illustration of the Learngene Condensing. We use the same task to train an ancestry model and pseudo descendant model. Then, we introduce a meta-network to calculate the layer similarity score between these two models. Finally, we select the most relevant model layers as the learngene by considering those that surpass the mean probability. Besides, it is straightforward to extend this process to the pseudo descendant models trained in different settings.}
\label{fig:paradigm} 
\end{center}
\end{figure*}

In this paper, we propose preliminary solutions to these three issues. Specifically, we set the learngene to some integral layers since prior studies~\cite{selvaraju2017grad, jiang2021layercam} show that integral layers can preserve the significant knowledge. In this way, condensing is simplified to choose suitable layers from the ancestry model, which is achieved by using a technique similar to a meta-learning mechanism~\cite{finn2017model, vanschoren2018meta}. As shown in Figure~\ref{fig:paradigm}, we create a pseudo descendant model to help choose layers from the ancestry model. Specifically, the layers from the ancestry model that consistently yield the most similar outputs to some layers of the descendant models across different scenarios are identified and treated as the extracted learngene. These scenarios might involve different training tasks or diverse random initializations, for example. After extracting the learngene layers with well-trained parameters, we move to the inheriting stage where we directly stack certain randomly initialized layers onto the learngene layers. This is done to construct diverse descendant models for addressing specific downstream tasks.

The remainder of the paper is organized as follows. First, we review the related work in Section~\ref{related_work}. Then we formulate $\textit{Learngene}$ in Section~\ref{sec:formulation_learngene} and elaborate on the technical details used to implement $\textit{Learngene}$ in both CNN and Transformer cases in Section~\ref{Methodology}. The experiments are carried out and the results are reported in Section~\ref{experiment}. Finally, Section~\ref{summary} summarizes the paper and discusses the future research directions.   

Our contributions are summarized as follows: 1) A novel machine learning paradigm named \textit{Learngene} is proposed and formulated; 2) The key technical issues in \textit{Learngene} are further enumerated; 3) For each key issue, the specialized solution is presented; 4) Through comprehensive experimentation, we substantiate four pivotal advantages of \textit{Learngene} and provide two additional analysis; 5) We discuss the possible application scenarios of \textit{Learngene} and future directions. 
Compared with our preliminary work~\cite{wang2021learngene}, this paper extends beyond it, offering the following significant improvements. 1) We propose an algorithm to automatically extract the learngene instead of the heuristic strategy used in~\cite{wang2021learngene}. 2) Besides the VGG structure used in~\cite{wang2021learngene}, \textit{Learngene} is extended to two more popular architectures which are ResNet~\cite{he2016deep} and ViT~\cite{dosovitskiy2021an}. 3) More experiments are conducted and discussed in greater detail.

\section{Related Work}
\label{related_work}

\textbf{Transfer Learning: }
Transfer learning~\cite{pratt1992discriminability, zhuang2020comprehensive, iman2022review} highlights transferring knowledge across domains and aims to improve the performance of target learners in the target domains. In traditional transfer learning, weights are transferred from a pre-trained network~\cite{he2019rethinking, zoph2020rethinking} that exactly matches the architecture of the new network. However, our paradigm diverges from this approach in that we do not need to transfer the original whole model. Instead, we condense an ancestry model into the learngene and use it to initialize the different descendant models that are much smaller. Indeed, a much smaller model can also learn to generalize as effectively as a more complex model through knowledge distillation~\cite{hinton2015distilling}, a common method of knowledge transformation. This approach typically uses the output probability distribution of the source model, in addition to the original labels for providing extra information to the target model. However, unlike this,
\textit{Learngene} employs some integral layers containing significant knowledge to initialize the target descendant models. Apart from the output predictions, the target model seeks to match other statistics from the source model such as intermediate feature representations~\cite{Romero2015FitNetsHF, zhang2021self}, Jacobian matrices~\cite{srinivas2018knowledge} and Gram matrices~\cite{srinivas2018knowledge}.  In fact, other layers in the descendant model in addition to the learngene
can transfer the intermediate feature representations from the ancestry model, thus reducing the gap of parameters between layers of the descendant model and stabilizing the features from layers of the descendant model. This allows the descendant model to better leverage its potential, especially in fewer samples scenarios.


\textbf{Parameter-Efficient Training: }
As the size of recent models continues to increase rapidly, it has become increasingly important to update models in a parameter-efficient manner. Recently, parameter-efficient training has been proposed that involves updating only newly added parameters (added either to the input or to the model)~\cite{houlsby2019parameter, Mahabadi2021ParameterefficientMF, karimi2021compacter, lester2021power, Li2021PrefixTuningOC}. In particular, adapters are widely used in computer vision~\cite{rebuffi2017learning, sung2022vl} and natural language processing~\cite{houlsby2019parameter, Mahabadi2021ParameterefficientMF, karimi2021compacter}. While adapters add the additional parameters to the models, prompt-based approaches add trainable parameters to the inputs~\cite{lester2021power, Li2021PrefixTuningOC}, and experiments have demonstrated their effectiveness. The main difference between \textit{Learngene} and parameter-efficient training is that \textit{Learngene} only reuses a more compact piece, while parameter-efficient training reuses the entire pre-training model. Therefore, \textit{Learngene} possesses stronger domain adaptation capabilities and results in lower costs when deploying its own models for the downstream tasks.


\textbf{Meta-learning and Bilevel Optimization: } 
Meta-learning, also known as learning-to-learn~\cite{schmidhuber1987evolutionary,  thrun2012learning, bengio2013optimization}, refers to the automatic process of model selection and algorithm tuning. Ordinarily, it has been extensively studied in the context of few-shot learning~\cite{finn2017model, snell2017prototypical, sung2018learning, wang2019few, liu2020negative, zhang2020deepemd, ye2020few, wang2021bridging, laenen2021episodes, fei2021melr}. Our aim focuses on extending meta-learning paradigms to solve the problem of transferring
 the meta-knowledge learned in the ancestry model to a practical lightweight model. Hence, this aim differs from the previous work in meta-learning for few-shot learning or domain generalization~\cite{li2018learning, dou2019domain, zhang2021adaptive, ojha2021few}. Besides, some meta-learning methods learn loss weights for the sample weights, such as discounting noisy samples~\cite{ren2018learning, shu2019meta, wang2020training, xu2021faster} and correcting class imbalance~\cite{shu2019meta, zhang2021learning}. In contrast, our paradigm not only 
 learns the loss weights but also extracts the learngene layers in the model by manipulating weights. Meta-learning methods are commonly formed as bilevel optimization problems~\cite{colson2007overview, sinha2017review, franceschi2018bilevel}, where an optimization problem contains another optimization problem as a constraint. Bilevel optimization is conducted on the top of an inner-level algorithm, tuning it to perform better on the target tasks. The meta-algorithm acts at an outer-level, based on the inner-level algorithm to compute a meta-objective and corresponding meta-gradient. Our work constructs a meta-objective as the regularization term to meet the bilevel optimization condition.

\textbf{Model Initialization and Inheritable Models: }
Model initialization can be categorized into weight and architecture initialization. Weight initialization involves setting the initial values for the weights of a neural network, which serve as a starting point for optimizing the model during training. Specifically, weight initialization usually transforms all of the parameters of a neural network at the beginning of learning the target tasks such as random initialization, Xavier initialization~\cite{glorot2010understanding}, Kaiming initialization~\cite{he2016deep}, self-distillation~\cite{furlanello2018born, zhang2021self}. On the other hand, architecture initialization refers to designing the structure of the neural network, which can be done either manually by a human expert or automatically using techniques such as Neural Architecture Search (NAS)~\cite{elsken2019neural}. However, learngene is a function that represents smaller architecture instead of a whole neural network and is then inherited to initialize the lightweight descendant models.  Recently, the concept of an inheritable model~\cite{kundu2020towards} is introduced for unsupervised domain adaptation. Different from the prior art, our work automatically seeks certain integral layers of the model as the inheritable layers for initializing the descendant model. 


\section{Formulation of learngene}
\label{sec:formulation_learngene}


For clarity, the primary notations used throughout this paper are listed in Table~\ref{table:summary_notation}. In general, the ancestry model $\mathcal{F}(\cdot;\bm{\Theta})$ is a complex pre-trained model (e.g, a ViT~\cite{dosovitskiy2021an} or CNN~\cite{DBLP:journals/corr/SimonyanZ14a, he2016deep} trained on the source data) or a model that continually learns from a stream of data~\cite{delange2021continual}. We define a function $\mathcal{G}(\cdot;\bm{\phi}):\bm{\Theta}  \rightarrow \bm{\theta}_{\mathbf{L}}$ that condenses the ancestry model into a more compact information piece named as learngene with the parameter $\bm{\theta}_{\mathbf{L}}$. Moreover, we set $|\bm{\theta}_{\mathbf{L}}|<|\bm{\Theta}|$ to highlight that the information of learngene is extracted from the ancestry model. 

The form of learngene can vary depending on the specific setting and the condensing function $\mathcal{G}$ should also change based on the form of the learngene. In this paper, we set learngene to certain integral layers of the ancestry model, which is naturally smaller than the whole model, and $\mathcal{G}$ is set to a meta-network that decides whether each layer of the ancestry model should be learngene or not.

In the inheriting stage, for each descendant model $f(\cdot;\bm{\theta})$, the initialized parameter $\bm{\theta}$ contains two parts: the condensed learngene $\bm{\theta}_{\mathbf{L}}$ and the randomly initialized $\bm{\theta}_{0}$:
\begin{equation}
\label{eq:hat_theta_0}
\bm{\theta} = \bm{\theta}_{0} \circ \bm{\theta}_{\mathbf{L}} ,
\end{equation}
In this paper, since we treat learngene as the integral layers, we stack certain randomly initialized layers parameterized by $\bm{\theta}_{0}$ to the learngene layers to construct the descendant model, which will be trained by the corresponding downstream task for quick adaptation. 


\begin{table}
\caption{Summary of the Mainly Used Notations.}
\label{table:summary_notation}
\begin{scriptsize}
\begin{tabular}{ll}
\hline
\toprule[1pt]
Symbol                                             & Definition                                             \\ \hline
$\mathcal{F}(\cdot;\bm{\Theta})$                  & ancestry model                                       \\
$f(\cdot;\bm{\theta})$                            & descendant model                                       \\
$\bm{\Theta}$                             & ancestry model's parameter                     \\
$\bm{\theta}$                            & descendant model's parameter                      \\
$\bm{\theta}_{\mathbf{L}}$                           & learngene parameter                             \\
$\bm{\theta}_{0}$                           & randomly initialized parameter of the descendant model         \\
$\mathcal{F}^l(\cdot;\bm{\Theta}^l)$   & the $l$-th layer of the ancestry model\\
$\bm{Z}^l$   & the feature output from $\mathcal{F}^l(\cdot;\bm{\Theta}^l)$\\
$f^k(\cdot;\bm{\theta}^k)$   & the $k$-th layer of the descendant model\\
$\bm{z}^k$   & the feature output from $f^k(\cdot;\bm{\theta}^k)$\\
$L$   & the total number of layers in the ancestry model\\
$K$   & the total number of layers in the descendant model\\
$\alpha^{l,k}$                       & layer similarity score between $\mathcal{F}^l(\cdot;\bm{\Theta}^l)$ and $f^k(\cdot;\bm{\theta}^k)$  \\
$\bm{\psi}^{l,k}$ & feature similarity matrix between $\bm{Z}^l$ and  $\bm{z}^k$\\
$\mathcal{G}(\cdot;\bm{\phi})$  & the meta-network to extract learngene \\ 
$h(\cdot;\bm{\zeta})$ & the alignment function \\
\bottomrule[1pt]
\hline
\end{tabular}
\end{scriptsize}
\end{table}

\section{Methodology}
\label{Methodology}
In this section, we propose preliminary solutions for the above-mentioned three key issues of \textit{Learngene}: (\rmnum{1}) learngene form, (\rmnum{2}) condensing function, and (\rmnum{3}) inheriting process. First, the form of learngene is introduced in Section~\ref{method:species_learngene} and is further applied to mainstream network architectures, \eg ViT and CNNs. Then we adopt a mechanism similar to a meta-learning method~\cite{finn2017model,vanschoren2018meta} to automatically extract learngene and derive a theoretically-sound convergence rate in Section~\ref{method:extraction_learngene}. Finally, the learngene that preserves significance is inherited to initialize the descendant models (cf. Section~\ref{method:initialization_individual}).




\subsection{Form of Learngene}
\label{method:species_learngene}
In Section~\ref{sec:formulation_learngene}, we define learngene as a more compact piece as long as it can condense significant knowledge from an ancestry model while we do not specify what form it can be. For a deep network, an integral layer is one elementary building block and more importantly, some previous studies~\cite{selvaraju2017grad, jiang2021layercam} qualitatively demonstrate that some integral layers contain significant knowledge (\eg the local texture or semantic concept). Considering these two characteristics, we set the integral layers of an ancestry model as the learngene. For the ancestry model, we choose the two most widely used network architectures as the case studies, which are ViT~\cite{dosovitskiy2021an} and CNNs (including VGG~\cite{DBLP:journals/corr/SimonyanZ14a} and ResNet~\cite{he2016deep}).

Mathematically, let $\mathcal{F}^l(\cdot;\bm{\Theta}^l)$ denotes the $l$-th integral layer of the ancestry model $\mathcal{F}(\cdot;\bm{\Theta})$ and the corresponding output feature can be formulated as a recursive update:
\begin{equation}
\label{eq:feature_collective_layer_l}
\bm{Z}^l=\mathcal{F}^l(\bm{Z}^{l-1};\bm{\Theta}^l), \quad l=1, \ldots, L ,
\end{equation}
where $\bm{Z}^l$ denotes the feature embedding in layer $l$, and $L$ is the total number of layers in the ancestry model. Next, we introduce the specific form of learngene in ViT and CNNs.


\textbf{The form of learngene in ViT.} 
If $\mathcal{F}(\cdot;\bm{\Theta})$ is a ViT, $\mathcal{F}^l(\cdot;\bm{\Theta}^l)$ includes multi-head self-attention (MSA), LayerNorm (LN), and Feed-Forward Network (FFN) layers.
Thus, Eq.~\eqref{eq:feature_collective_layer_l} in ViT becomes:
\begin{equation}
\label{eq:ViT-encoder}
\begin{aligned}
&\hat{\bm{Z}}^l=\mathrm{MSA}\left(\operatorname{LN}\left(\bm{Z}^{l-1}\right)\right)+\bm{Z}^{l-1}, \\
&\bm{Z}^l=\operatorname{FFN}\left(\operatorname{LN}\left(\hat{\bm{Z}}^l\right)\right)+\hat{\bm{Z}}^l ,
\end{aligned}
\end{equation}
where $\hat{\bm{Z}}^l$ and $\bm{Z}^l$ represent the output features of the MSA module and the FFN module for $l$-th block, respectively. Here, $\bm{Z}^l$ has dimensions $\mathbb{R}^{P \times D}$, where $P$ and $D$ denote the number of patches and dimensions of tokens, respectively. Each element of the feature $\bm{Z}^l$ is indexed as $\bm{Z}^l_{i, j}$,  where $i \in\{1,2, \ldots P\}$ and $j \in$ $\{1,2, \ldots, D\}$. 

\textbf{The form of learngene in CNNs.}
If $\mathcal{F}(\cdot;\bm{\Theta})$ is a CNN, \eg VGG~\cite{DBLP:journals/corr/SimonyanZ14a}, $\mathcal{F}^l(\cdot;\bm{\Theta}^l)$ usually includes BatchNorm (BN) and Convolutional (Conv) layers. In this case, Eq.~\eqref{eq:feature_collective_layer_l} becomes:
\begin{equation}
\label{eq:vgg_layer}
\bm{Z}^l=\operatorname{BN}\left(\mathrm{Conv}\left(\bm{Z}^{l-1}\right)\right),
\end{equation}
where $\bm{Z}^l \in \mathbb{R}^{C \times H \times W}$ and $[C;H;W]$ represents the channel size, height, and width, respectively. The element of the feature $\bm{Z}^l$ is indexed as $\bm{Z}^l_{c, i, j}$ where $c \in\{1,2, \ldots C\}$, $i \in\{1,2, \ldots,  H\}$ and $j \in$ $\{1,2, \ldots, W\}$. Note that we do not use Fully Connected (FC) Layers as the learngene since these layers can be replaced by global average pooling~\cite{lin2013network} and the size of the corresponding parameters is redundant. For ResNet~\cite{he2016deep}, which incorporates skip connections,  Eq.~\eqref{eq:feature_collective_layer_l} changes into:
\begin{equation}
\label{eq:resnet_layer}
\bm{Z}^l=\operatorname{BN}\left(\mathrm{Conv}\left(\bm{Z}^{l-1}\right)\right)+\bm{Z}^{l-1},
\end{equation}



\subsection{Learngene Condensing}
\label{method:extraction_learngene}
In this section, we answer the second issue that how to condense an ancestry model into the learngene. After designating learngene as the integral layer, we streamline the condensation process by choosing suitable layers from the ancestry model. To accomplish this, we employ a mechanism akin to a meta-learning technique~\cite{finn2017model,vanschoren2018meta}, which is detailed in Section~\ref{method:extraction_process}. Then we introduce how to optimize it in Section~\ref{method:optimizing_scheme} and also derive the theoretical analysis of the optimization scheme in Section~\ref{method:convergence_analysis}. 

\subsubsection{Condensation Process}
\label{method:extraction_process}
In the last section, we show that treating integral layers as the learngene is reasonable, then we may wonder how to choose suitable layers from the ancestry model to preserve significance
 (\ie realizing the function $\mathcal{G}(\cdot;\bm{\phi})$ in Section~\ref{sec:formulation_learngene}). To achieve this goal, we use a technique similar to a meta-learning mechanism~\cite{finn2017model,vanschoren2018meta} which can learn an adaptive weighting scheme from data to make the choice more automatic and reliable.

To condense an ancestry model into the learngene, we first set up a pseudo descendant model. Then we identify which layers in the ancestry model have the most similar outputs as some layers of the pseudo descendant models trained by the different settings, \eg different tasks or initializations. These layers are extracted as learngene layers because they are able to retain the most significant knowledge across different tasks and consistently produce outputs similar to those of the respective layers in the pseudo descendant models, regardless of the initialization. 

Let $f^k(\cdot;\bm{\theta}^k)$ denotes the $k$-th layer of the pseudo descendant model $f(\cdot;\bm{\theta})$ and $\bm{z}^k$ denotes the corresponding output. This can be expressed as:
\begin{equation}
\label{eq:feature_individual_layer_k}
\bm{z}^{k}=f^k(\bm{z}^{k-1};\bm{\theta}^k), \quad k=1, \ldots, K ,
\end{equation}
where $K$ is the total number of layers in $f(\cdot;\bm{\theta})$. 

We first show how to calculate the feature similarities for a pseudo descendant model. This calculation is straightforward to extend to the pseudo descendant models trained in different settings. For ease of notation, we use $\bm{\psi}^{l,k}$ to denote the feature similarity matrix between each pair of the outputs $\bm{Z}^l$ and  $\bm{z}^k$, which is defined as follows:
\begin{equation}
\label{eq:similarity}
\bm{\psi}^{l,k} = \left(h(\bm{z}^k;\bm{\zeta})-\bm{Z}^l \right)^{2}. 
\end{equation}
where $h(\cdot;\bm{\zeta})$ is an alignment function which is proposed to handle the potential mismatch in feature dimensions between the outputs $\bm{Z}^l$ and $\bm{z}^k$, \eg using an identity mapping for ViT or pointwise convolution for CNNs. Given a $L$-layer $\mathcal{F}(\cdot;\bm{\Theta})$ and a $K$-layer $f(\cdot;\bm{\theta})$, we need to calculate $L \times K$ feature similarity matrix. Figure~\ref{fig:paradigm} shows one case where $L=5$ and $K=3$.


We average the previous pairwise feature matrix $\bm{\psi}^{l,k}$ across the feature dimensions to generate the pairwise feature similarity, denoted as $\bar{\psi}^{l,k}$. Next, we introduce a learnable parameter $\alpha^{l, k}$ for each pair (l, k) and weight it on $\bar{\psi}^{l,k}$ as a regularization term loss that needs to be optimized. In this way, $\alpha^{l, k}$ has a strong correlation with $\bar{\psi}^{l,k}$. For example, a large value of $\bar{\psi}^{l,k}$ indicates that the pair (l, k) of features are dissimilar, so $\alpha^{l, k}$ will be small, and vice versa. Therefore, $\alpha^{l, k}$ can be used as the pairwise layer similarities, which can serve as the basis for selecting which layers are learngene layers. Specifically, we set $\alpha^{l, k}=\mathcal{G}^{l, k}(\bm{Z}^l;\bm{\phi})$. The objective function for optimizing the parameter $\bm{\phi}$ of $\mathcal{G}(\cdot)$ is denoted as:

\begin{equation}
\label{eq:metaloss}
\mathcal{L}^{\mathrm{\text{meta}}}=\sum_{(l, k) \in R} \alpha^{l, k} \bar{\psi}^{l,k},
\end{equation}
where $R$ stands for a set of candidate pairs. We will elaborate on the optimization scheme for the meta-network that generates the layer similarity score $\bm{\alpha}$ in Section~\ref{method:optimizing_scheme}. 

After the optimization of the meta-network, we compute the maximum value of $\alpha^{l,k}$ over $k$ for each $l$, \ie $\alpha^{l}=\max(\alpha^{l,1}, \ldots, \alpha^{l,K})$. We then normalize $\alpha^{l}$ to have unit variance:
\begin{equation}
\label{eq:alphasoftmax}
\Tilde{\alpha}^{l} = \frac{\exp(\alpha^{l}) }{\sum_{j=1}^{L} \exp (\alpha^{j})},
\end{equation}
which measures how the $l$-th layer of the ancestry model is similar to the layers of the pseudo descendant model. Finally, we select the most relevant model layers as the learngene: 
\begin{equation}
\label{eq:integrate}
\bm{\theta}_{\mathbf{L}} \leftarrow \begin{cases} \text{return the }l\text{-th layer} & \text { if } \Tilde{\alpha}^{l} > \frac{1}{L} \\  \text{null} & \text { otherwise. }\end{cases},
\end{equation}
where $l \in\{1,2, \ldots, L\}$. As illustrated in Figure \ref{fig:paradigm}, the meta-network produces similarity scores for each layer in the ancestry model, where $\Tilde{\bm{\alpha}}^{1:5}=\{0.25, 0.15, 0.13, 0.16, 0.31\}$. Since $\Tilde{\alpha}^{1}$ and $\Tilde{\alpha}^{5}$ are higher than 1/5, the 1-st and 5-th layers of the ancestry model are selected as the learngene layers, which will be used to initialize the descendant model. Additionally, we repeat the process of learngene condensing by randomly initializing the pseudo descendant model 10 times and find that the result is consistent across all trials. Next, we describe how to apply this method in ViT and CNN cases.


\noindent\textbf{ViT Case. }   If  $\mathcal{F}(\cdot;\bm{\Theta})$ is a ViT, the dimension of pairwise tokens from the ancestry model and descendant model are consistent. Therefore, we can directly use the identity mapping and thus Eq.~\eqref{eq:similarity} in ViT becomes:

\begin{equation}
\label{eq:matching}
\psi^{l,k}_{i,j} = \left(z^{k}_{i,j}-Z^l_{i, j} \right)^{2}.
\end{equation}
where $\bm{\psi}^{l,k} \in \mathbb{R}^{P \times D}$, $i \in\{1,2, \ldots P\}$ and $j \in$ $\{1,2, \ldots, D\}$. 

Then, the layer similarity score $\bm{\alpha}$ optimized by the meta loss becomes:
\begin{equation}
\label{eq:metaloss_vit}
\mathcal{L}^{\mathrm{\text{meta}}}=\sum_{(l, k) \in R} \alpha^{l, k} \frac{1}{P D} \sum_{i, j} \psi^{l,k}_{i,j}.
\end{equation}
In section~\ref{exp:why_work}, we show that the meta-network selects the lower layers in ViT as the learngene. More importantly, some preceding research~\cite{raghu2021do, park2022how} has shown that the lower layers in ViT contain the local texture and semantic concept, \ie the significant knowledge.

\noindent\textbf{CNN Case.} If  $\mathcal{F}(\cdot;\bm{\Theta})$ is a CNN, the pairwise features of the ancestry model and descendant model may have inconsistent dimensions. In such case, we can use a pointwise convolution $h_{conv}(\cdot;\bm{\zeta})$ to align $\bm{z}^{k}$ with $\bm{Z}^l$ and then Eq.~\eqref{eq:similarity} becomes:

\begin{equation}
\label{eq:matching}
\psi^{l,k}_{c,i,j} = \left(h_{conv}(\bm{z}^{k};\bm{\zeta})_{c,i,j}-Z^l_{c, i, j} \right)^{2}. 
\end{equation}
 Similar to the size of features in the CNN, $\bm{\psi}^{l,k} \in \mathbb{R}^{C \times H \times W}$,  $c \in\{1,2, \ldots C\}$, $i \in\{1,2, \ldots,  H\}$ and $j \in$ $\{1,2, \ldots, W\}$.

As a result, the meta loss for learning the layer similarity score $\bm{\alpha}$ is modified to:

\begin{equation}
\label{eq:metaloss_cnn}
\mathcal{L}^{\mathrm{\text{meta}}}=\sum_{(l, k) \in R} \alpha^{l, k} \frac{1}{C H W} \sum_{c, i, j} \psi^{l,k}_{c,i,j}.
\end{equation}
Likewise, we demonstrate the effectiveness of the meta-network in extract the lower and deeper layers of CNNs as the learngene in section~\ref{exp:why_work}. Additionally, we provide insights into the role of the lower and deeper layers, where the lower layer is more sensitive to local texture, while the deeper layer focuses more on the semantic concept. The local texture and semantic concept are usually the significant knowledge~\cite{selvaraju2017grad, GilpinBYBSK18}.


\subsubsection{Optimizing Scheme}
\label{method:optimizing_scheme}
The optimizing process of learngene condensing can be divided into two parts: (1) updating the pseudo descendant model $f(\cdot;\bm{\theta})$ and the alignment function $h(\cdot;\bm{\zeta})$ on the training data $\mathcal{D}$; (2) updating the parameter $\bm{\phi}$ of a meta-network $\mathcal{G}(\cdot)$ on the meta-data $\widehat{\mathcal{D}}$. Notably, the training data $\mathcal{D}$ and the meta-data $\widehat{\mathcal{D}}$ do not share any data points (\ie $\mathcal{D} \cap \widehat{\mathcal{D}}=\varnothing$), and they are both derived from the validation set.\footnote{The  detail is given in Appendix A.1} 
Several previous algorithms~\cite{Romero2015FitNetsHF, Zagoruyko2017PayingMA, jang2019learning, park2019relational, murugesan2022auto} have focused on transferring feature information to maximize the performance of the target model by utilizing the similarity of pairwise features in Eq.~\eqref{eq:similarity}. However, these methods require computing similarity scores for the entire dataset, resulting in significant computational and storage burdens. In contrast, our method only uses the validation set to learn the similarity of the pairwise features, and its goal is to find suitable learngene layers from the ancestry model.

Therefore, the total loss $\mathcal{L}^{\text{total}}$ that trains the pseudo descendant model $f(\cdot;\bm{\theta})$ and the alignment function $h(\cdot;\bm{\zeta})$ takes the form:

\begin{equation}
\label{eq:totalloss}
\mathcal{L}^{\text{total}}=\mathcal{L}^{\mathrm{cls}}+\mathcal{L}^{\text{meta}} ,
\end{equation}
where $\mathcal{L}^{\mathrm{cls}}$ is the loss function for classification (\eg cross entropy loss function) and is
computed on the training data $\mathcal{D}$. Afterwards, we leverage gradient descent to optimize $f(\cdot;\bm{\theta})$ and $h(\cdot;\bm{\zeta})$.


After optimizing $f(\cdot;\bm{\theta})$ and $h(\cdot;\bm{\zeta})$ for a single iteration, we proceed to update the parameter $\bm{\phi}$ of a meta-network $\mathcal{G}(\cdot)$. The parameter $\bm{\phi}$ can be updated guided by the optimization objective, \ie moving the current parameter $\bm{\phi}$ along the objective gradient:
 

 \begin{equation}
\label{eq:updatephi}
\bm{\phi} \leftarrow \bm{\phi}-\hat{\beta} \nabla_{\bm{\phi}}  \frac{1}{M} \sum_{i=1}^{M} \mathcal{L}_{i}^{\text {meta }} ,
\end{equation}
where $\hat{\beta}$ is the learning rate and $M$ is the size of a batch $\widehat{\mathcal{B}}$ sampled from the meta-data set $\widehat{\mathcal{D}}$. The optimization algorithm can then be summarized in Algorithm \ref{alg:optimizing}. 

\begin{algorithm}[htb]
   \caption{Optimizing  meta-network}\label{alg:optimizing}

   {\bfseries Input:} \text { Training data } $\mathcal{D}$,\text { meta-data } $\widehat{\mathcal{D}}$ \text {, batch size } $N, M$, learning rate $\beta$, $\hat{\beta}$ , max iterations $I$\\
   {\bfseries Output:} Layer similarity score $\bm{\alpha}$ 
  \begin{algorithmic}[1]  
  \STATE Initialize the pseudo descendant model $f(\cdot;\bm{\theta})$ and the alignment function $h(\cdot;\bm{\zeta})$, and then initialize the parameter $\bm{\phi}$ of a meta-network.
   \FOR{$i=1,\ldots,I$}
   \STATE Sample a batch $\mathcal{B} \subset \mathcal{D}$ with $|\mathcal{B}|=N$, sample a batch $\widehat{\mathcal{B}} \subset \widehat{\mathcal{D}}$ with $|\widehat{\mathcal{B}}|=M$
   \STATE  Formulate the total loss function as Eq.~\eqref{eq:totalloss}
   \STATE $\bm{\theta} \leftarrow \bm{\theta}- \beta \nabla_{\bm{\theta}} \mathcal{L}^{\text{total}}$
   \STATE $\bm{\zeta} \leftarrow \bm{\zeta}- \beta \nabla_{\bm{\zeta}} \mathcal{L}^{\text{total}}$
   \STATE Update $\bm{\phi}$ by Eq.~\eqref{eq:updatephi}
   \ENDFOR
    \end{algorithmic}
\end{algorithm}


\subsubsection{Convergence Analysis of Optimization}
\label{method:convergence_analysis}

Here, our algorithm is a type of bi-level optimization scheme. 
We present the theoretical analysis for the convergence of meta-network.\footnote{The proof is given in Appendix~\ref{supp:theorem}} For the convenience of proof, we use $\bm{w}$ to collectively refer to the two parameters of $\bm{\theta}$ and $\bm{\zeta}$.\\
\textbf{Theorem 1. } \emph{Suppose the loss function $\mathcal{L}^{\text {meta}}$  satisfy Lipschitz smoothness and  the Hessian $\nabla^2\mathcal{L}^{\text {meta}}$ is $\rho$-Lipschitz continuous (\ie for every $u, v \in \mathbb{R}^{d}$, $\left\|\nabla^{2} \mathcal{L}^{\text {meta}}(u)-\nabla^{2} \mathcal{L}^{\text {meta}}(v)\right\| \leq \rho\|u-v\|$.). Gradient $\nabla \mathcal{L}^{\text {meta}}$ has a bounded variance w.r.t. meta-data or training data (\ie for any $z$, $\mathcal{B}$, $\mathbb{E}\|\nabla \mathcal{L}^{\text{meta}}(z; \mathcal{B})-\nabla \mathcal{L}^{\text{meta}}(z)\|^{2} \leq \sigma^{2}$ for some $\sigma>0$), and so is the Hessian $\nabla^2\mathcal{L}^{\text {meta}}$. The training function $L$ and the meta objective function $\mathcal{L}^{\text{meta}}$ are nonconvex w.r.t. $\bm{w}$ and $\bm{\phi}$. Let $\beta_w \in (0, \frac{1}{6 L}]$, and we have }
\begin{equation}
\label{eq:theorem-2}
\frac{1}{T} \sum_{t=0}^{T-1} \mathbb{E} \left[\left\|\nabla \mathcal{L}_{t}^{\text {meta}}(\bm{\phi} \mid \bm{w})\right\|^{2}\right] \leq \mathcal{O}(\frac{L_{\bm{\phi}}}{\sqrt{T}}),
\end{equation}
where $L_{\bm{\phi}}$ is some constant independent of the convergence process. This demonstrates that the meta-network can converge and stably learn the layer similarity $\alpha$.

\subsection{Inheriting Learngene}
\label{method:initialization_individual}

After extracting the learngene layers $\bm{\theta}_{\mathbf{L}}$ with well-trained parameters, we move on to the inheriting stage. In this stage, we directly stack certain randomly initialized layers $\bm{\theta}_{0}$ to the learngene layers $\bm{\theta}_{\mathbf{L}}$ to construct diverse descendant models for solving the specific downstream tasks. 
Next, we introduce how to initialize the specific networks.






\noindent\textbf{Initialize ViT. } 
Since the features output from a Transformer layer can be generally used as direct input to the next Transformer layer, we sequentially leverage the inherited learngene layers as the whole Transformer encoder and stack them to a classification head of the randomly initialized parameter for classification tasks. In addition, the pre-processing embeddings (\eg the patch embeddings, [class] token, and position embeddings) in ViT are also randomly initialized. 

\noindent\textbf{Initialize CNNs. } 
In order to change the size of features from large to small and enable effective classification, CNN requires convolutional layers and pooling operations in the middle. Therefore, we randomly initialize certain convolutional layers and one fully connected layer, and then stack them to the learngene layers to form a descendant model for classifying.

Moreover, the descendant model is smaller than the ancestry model. Since the condensed learngene preserves the significant knowledge of the big ancestry model, it makes descendant models quickly adapt to different tasks by a few training samples. Even if the descendant models encounter diverse data domains, \textit{Learngene} can relieve the domain-shift~\cite{luo2019taking}.

\begin{figure*}[htbp]
\setlength{\abovecaptionskip}{0pt}
\setlength{\belowcaptionskip}{0pt}
  \centering
  \subfigure[ViT, 10-shot ]{
    \label{fig:ViT-shot}
    \includegraphics[width=0.26\textwidth,trim=1 1 1 60,clip]{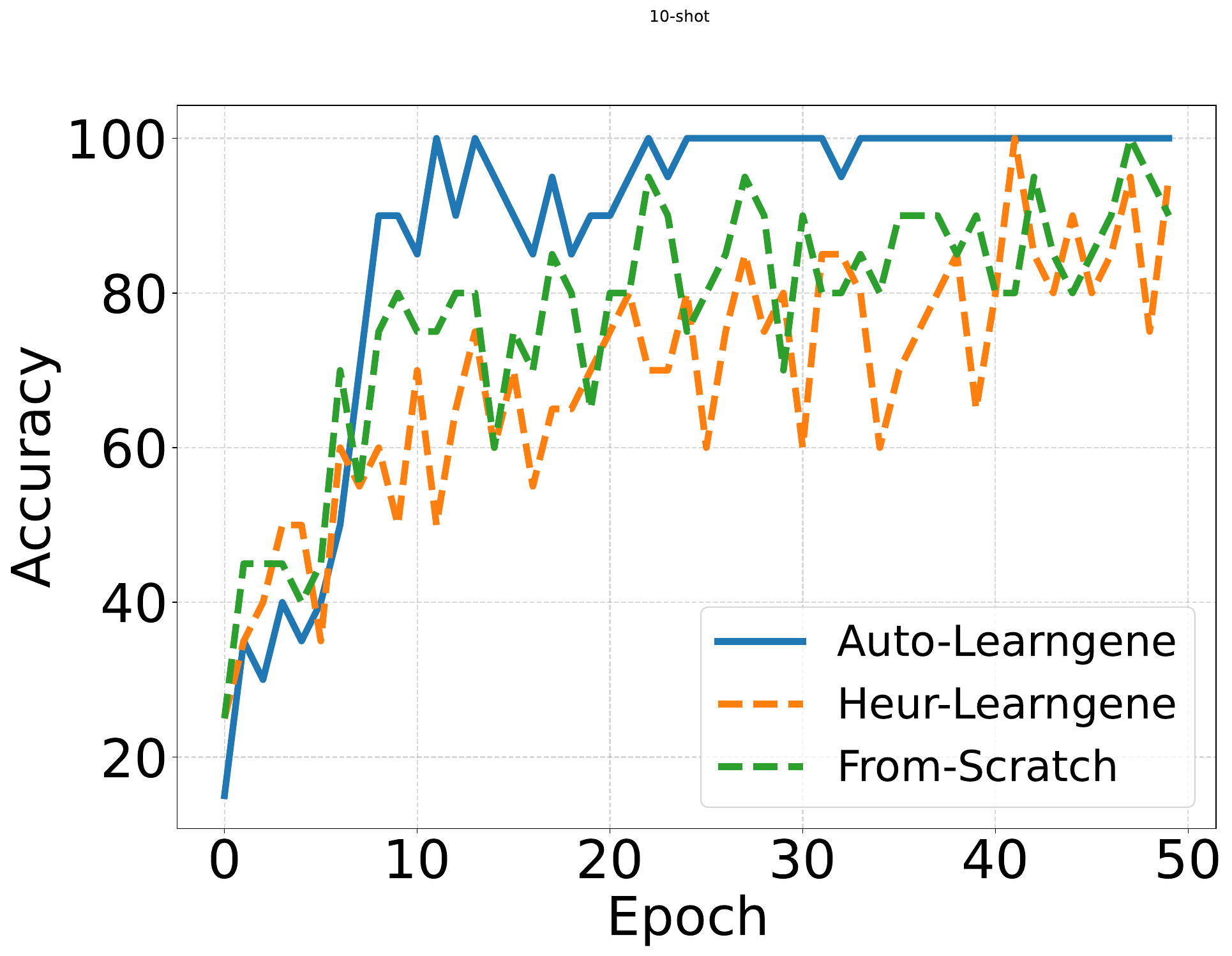}} 
    \subfigure[ViT, 20-shot ]{
    \label{fig:ViT20-shot}
    \includegraphics[width=0.26\textwidth,trim=1 1 1 60,clip]{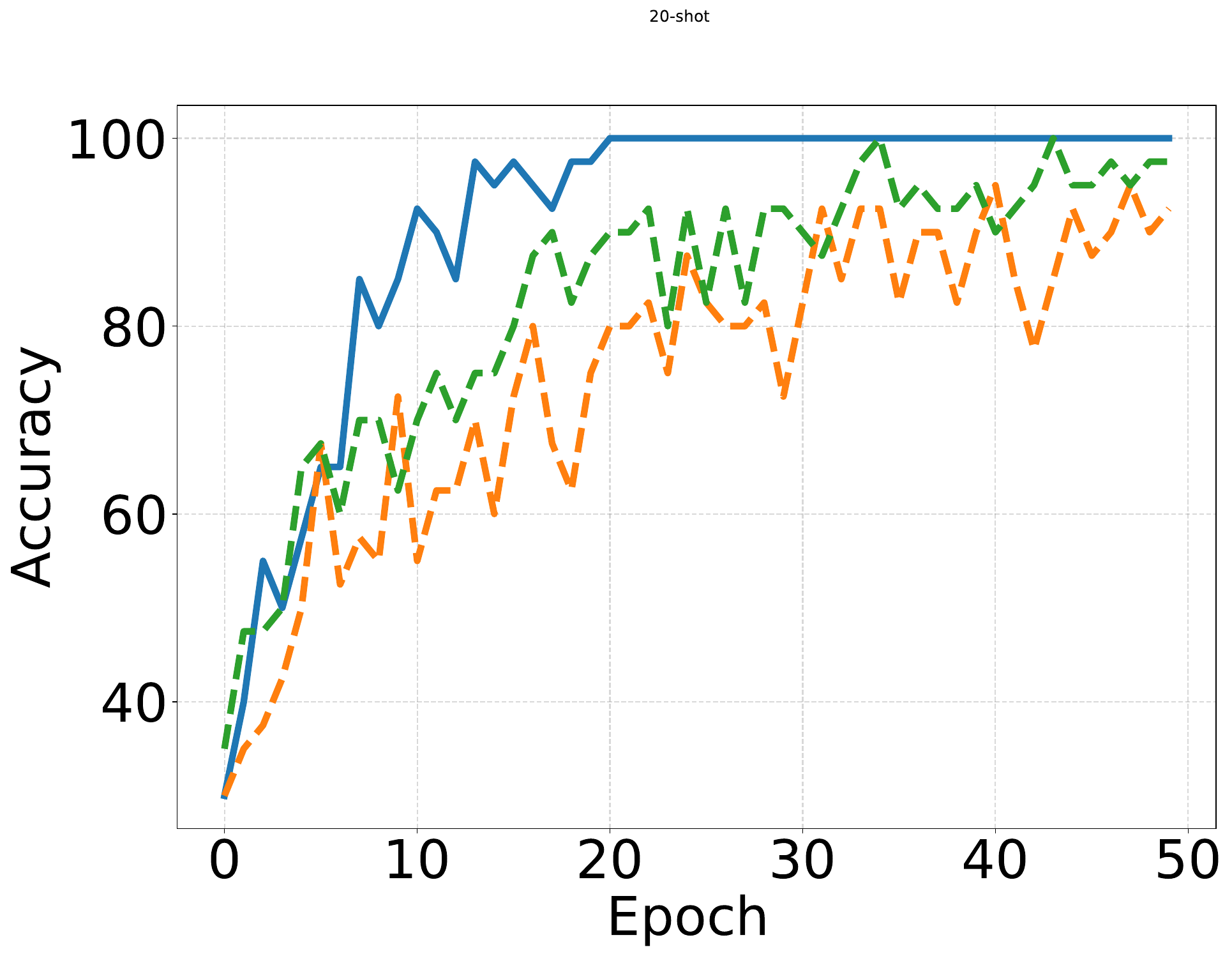}} 
    \subfigure[ViT, 30-shot]{
    \label{fig:ViT30-shot}
    \includegraphics[width=0.26\textwidth,trim=1 1 1 60,clip]{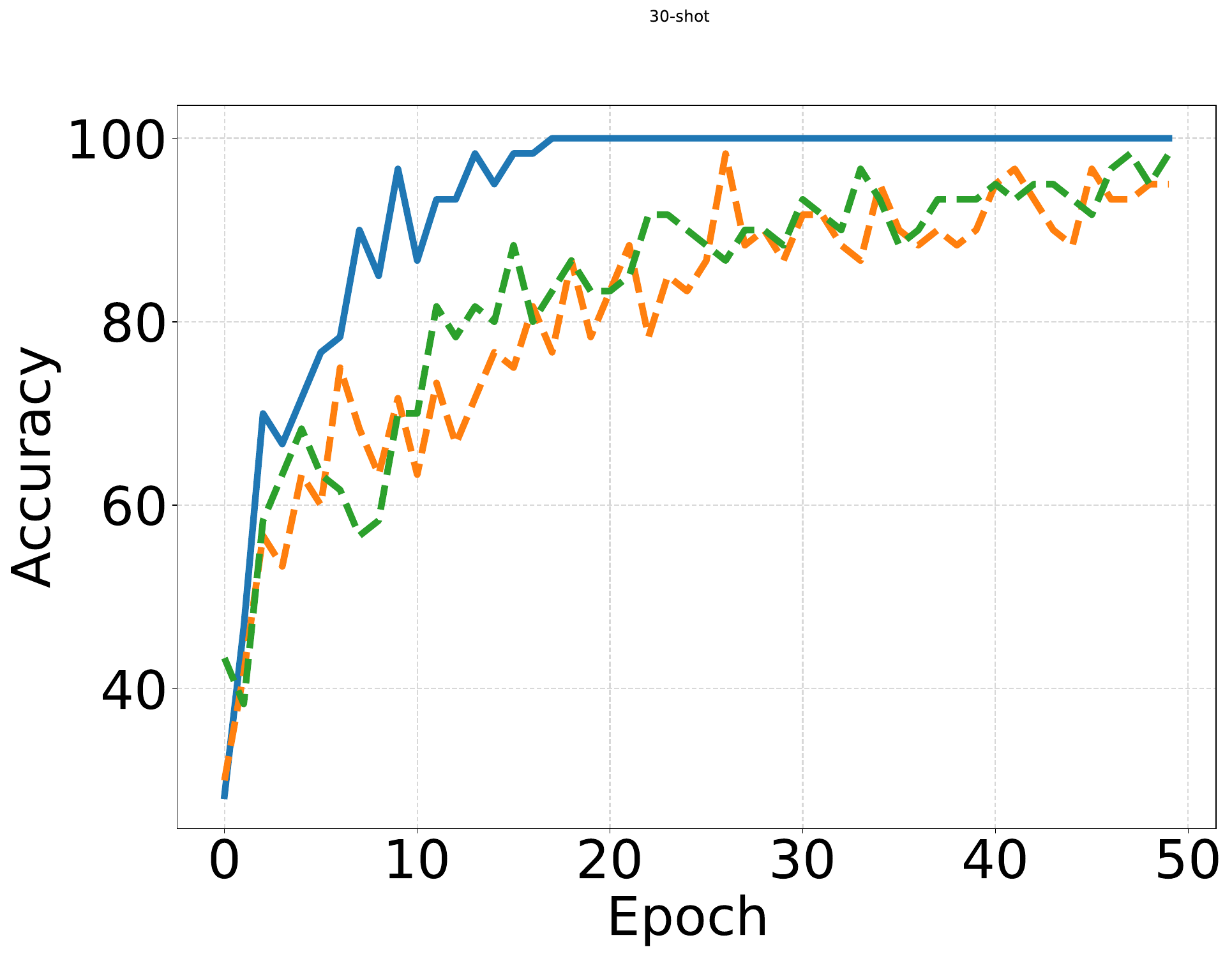}} 
    \subfigure[ViT, 40-shot]{
    \label{fig:ViT40-shot}
    \includegraphics[width=0.26\textwidth,trim=1 1 1 60,clip]{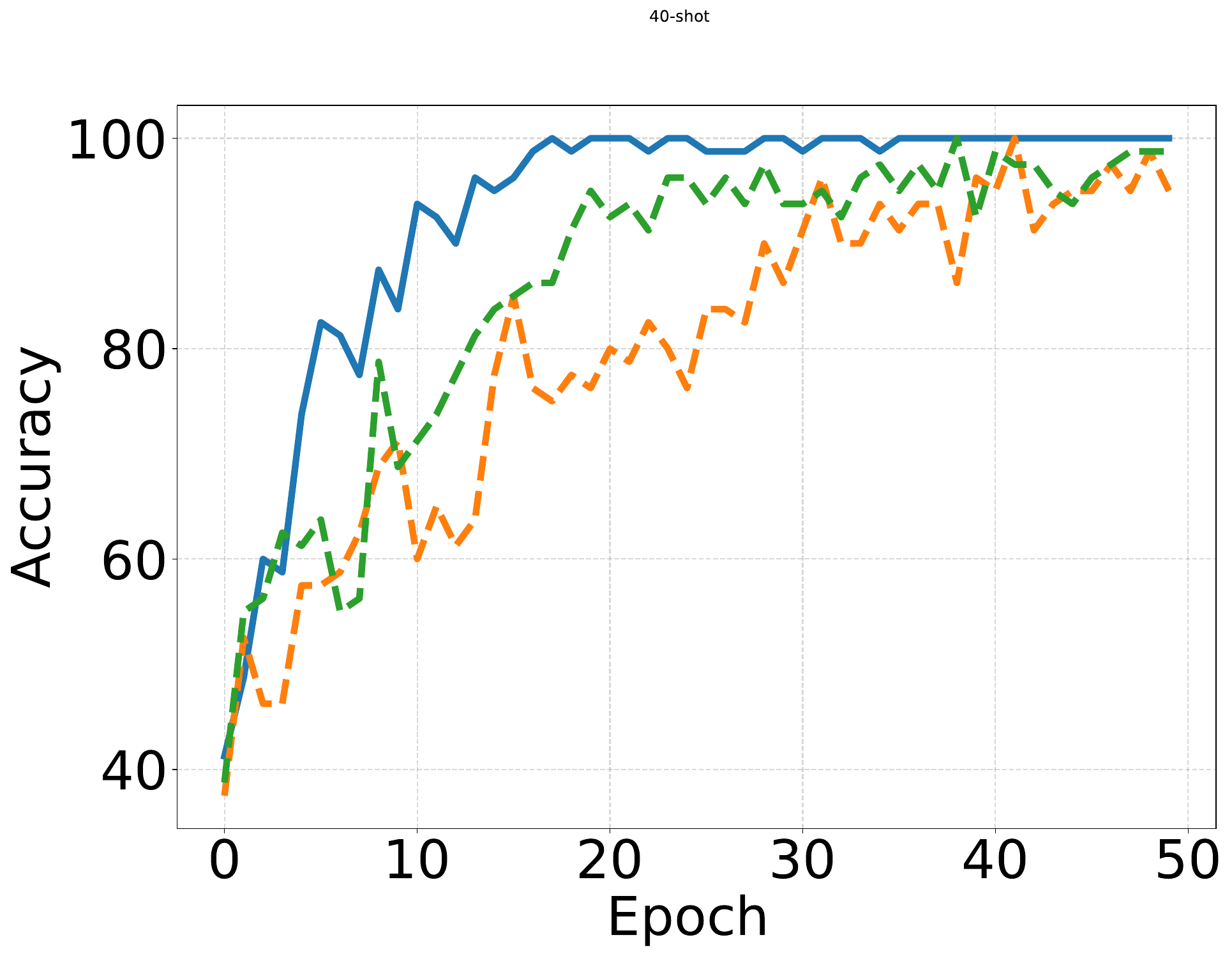}} 
    \subfigure[ViT, 50-shot]{
    \label{fig:ViT50-shot}
    \includegraphics[width=0.26\textwidth,trim=1 1 1 60,clip]{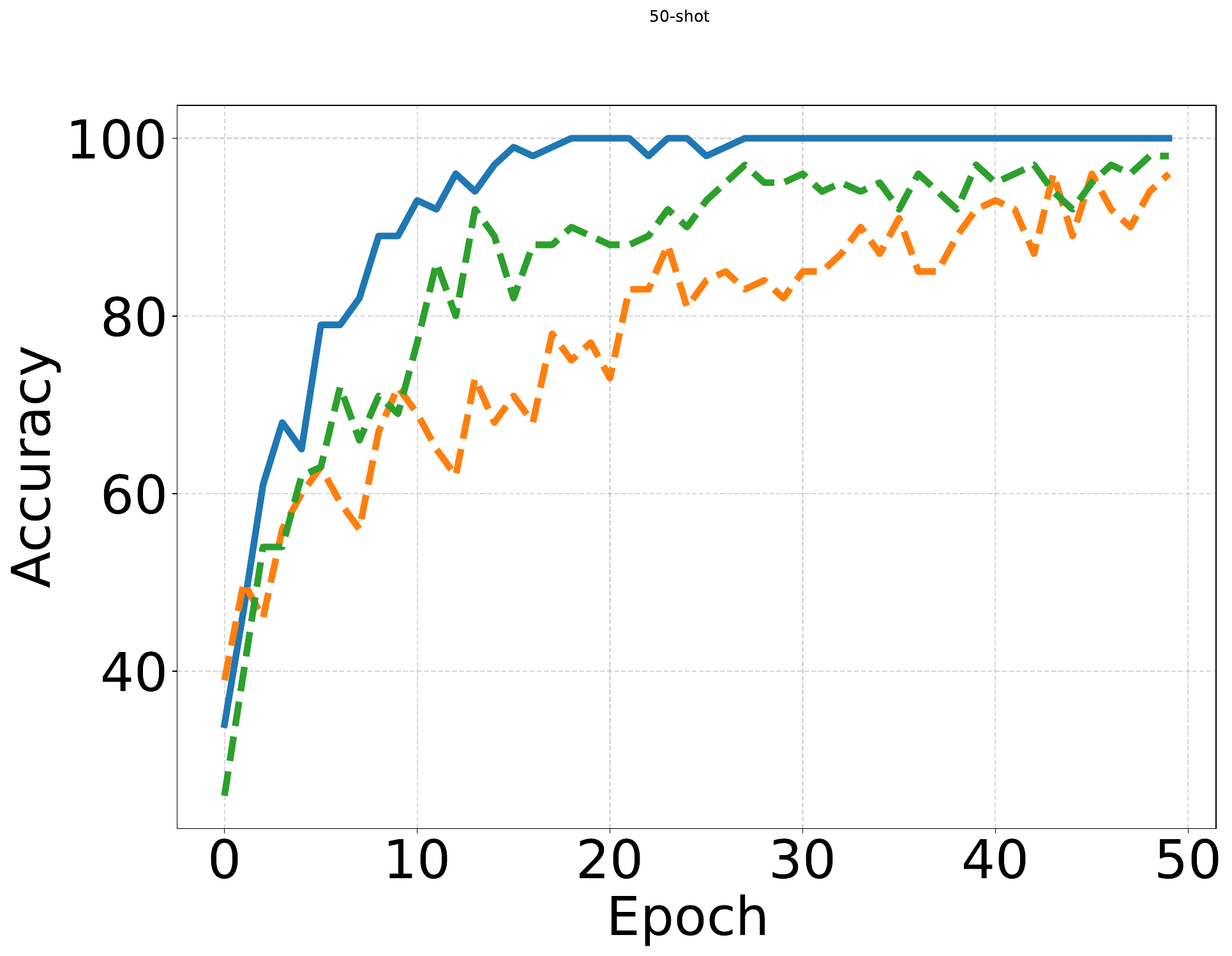}} 
    \subfigure[ViT, 60-shot]{
    \label{fig:ViT60-shot}
    \includegraphics[width=0.26\textwidth,trim=1 1 1 60,clip]{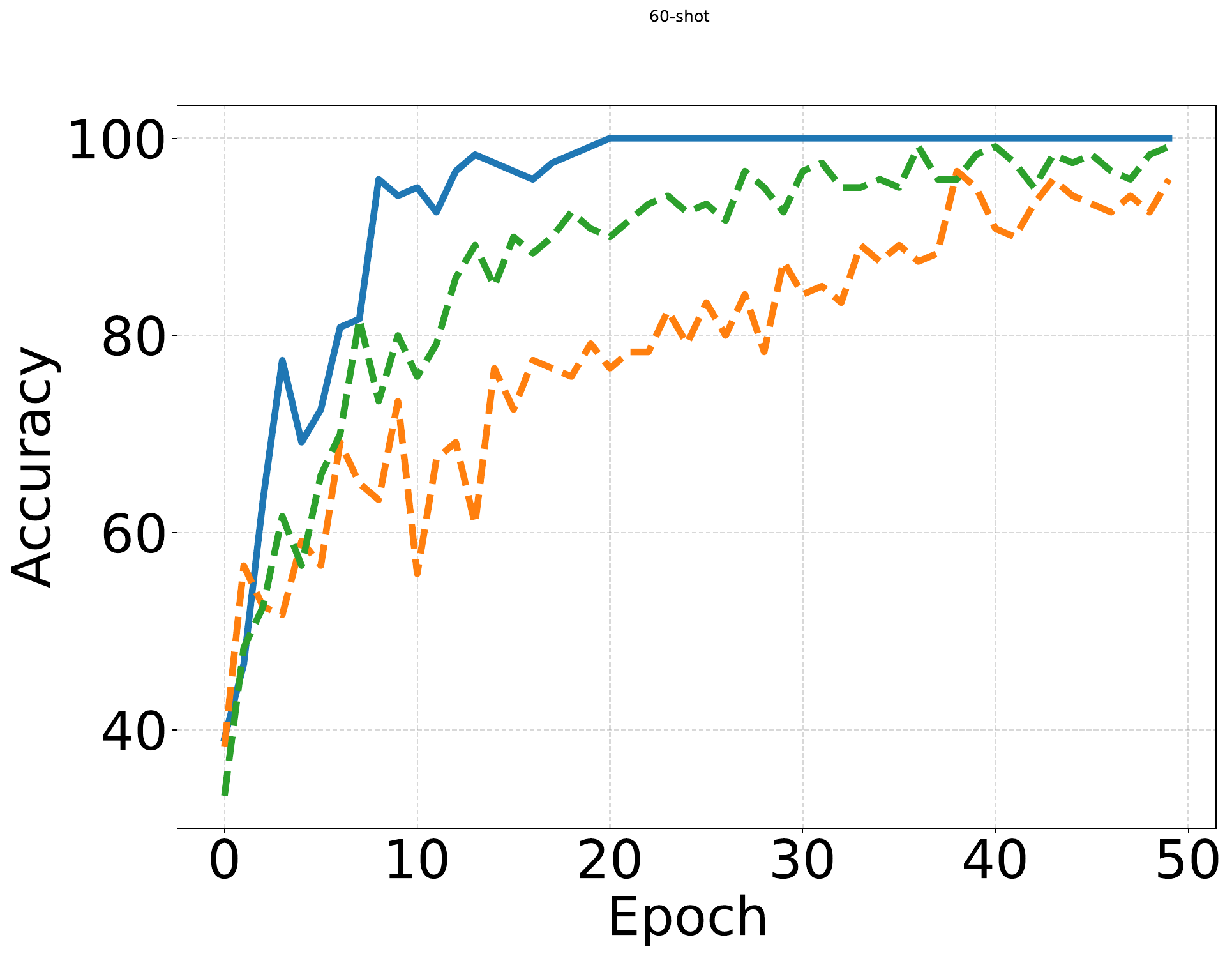}} 
    \subfigure[VGG, 10-shot ]{
    \label{fig:VGG10-shot}
    \includegraphics[width=0.26\textwidth,trim=1 1 1 60,clip]{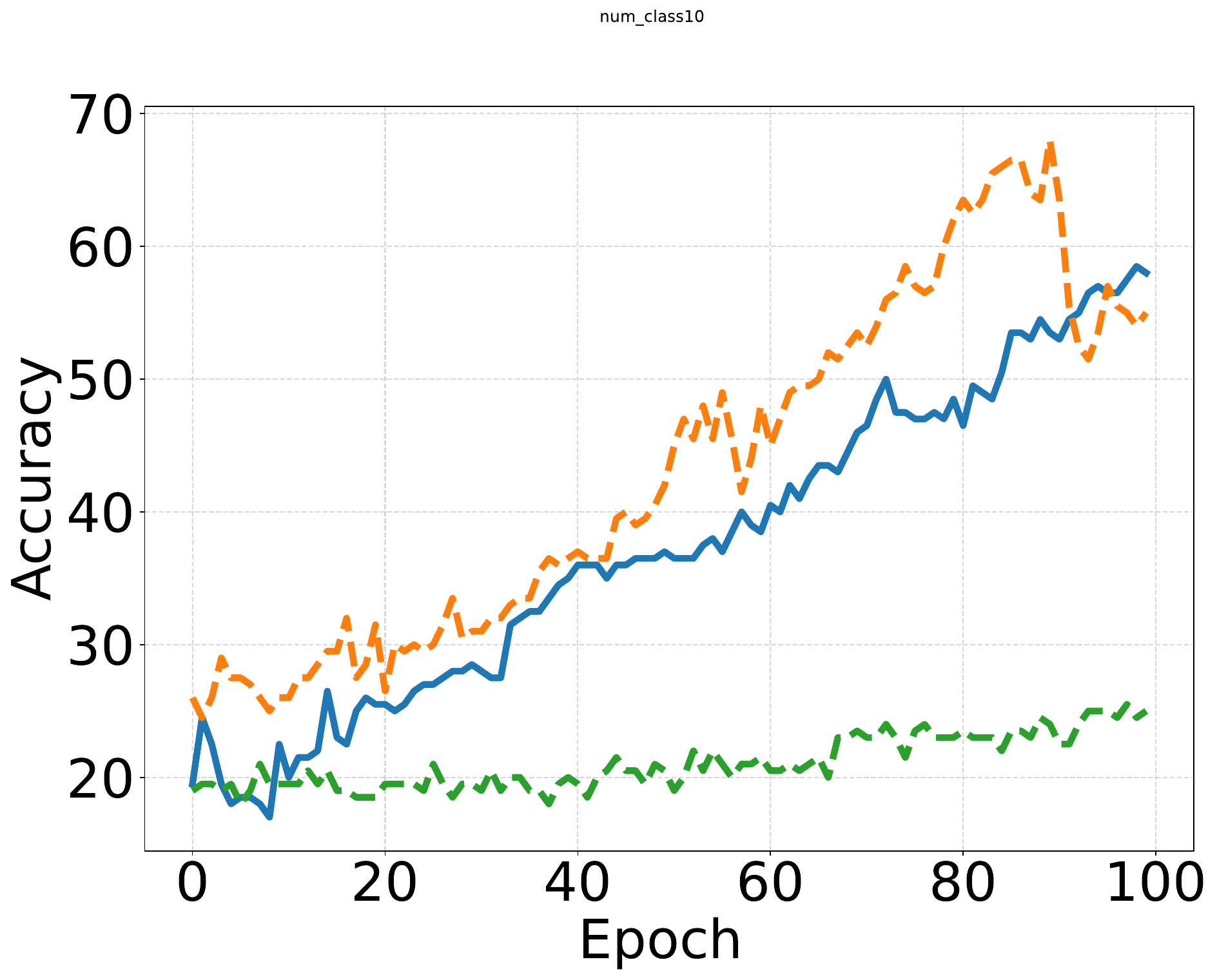}} 
    \subfigure[VGG, 20-shot ]{
    \label{fig:VGG20-shot}
    \includegraphics[width=0.26\textwidth,trim=1 1 1 60,clip]{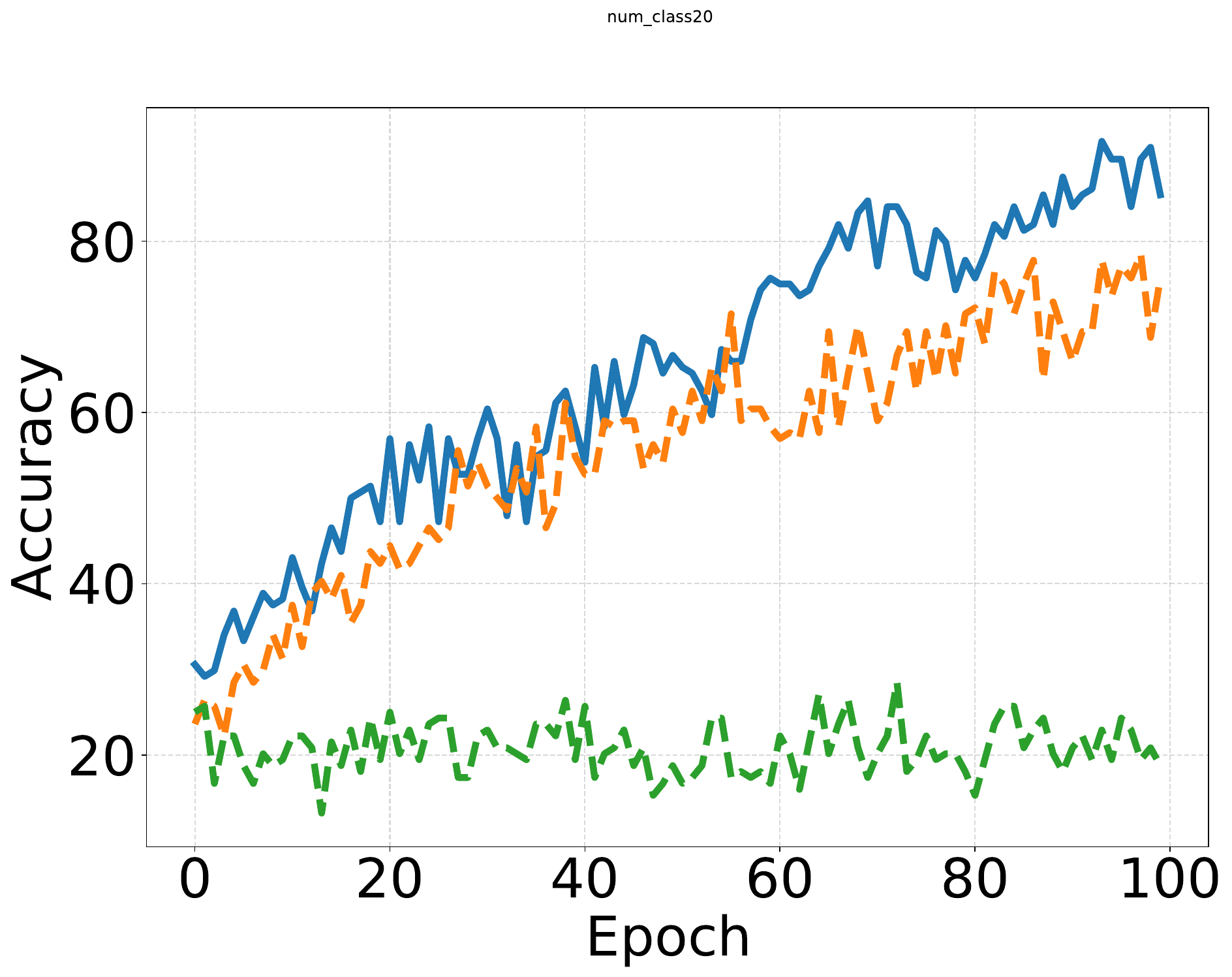}} 
    \subfigure[VGG, 30-shot]{
    \label{fig:VGG30-shot}
    \includegraphics[width=0.26\textwidth,trim=1 1 1 60,clip]{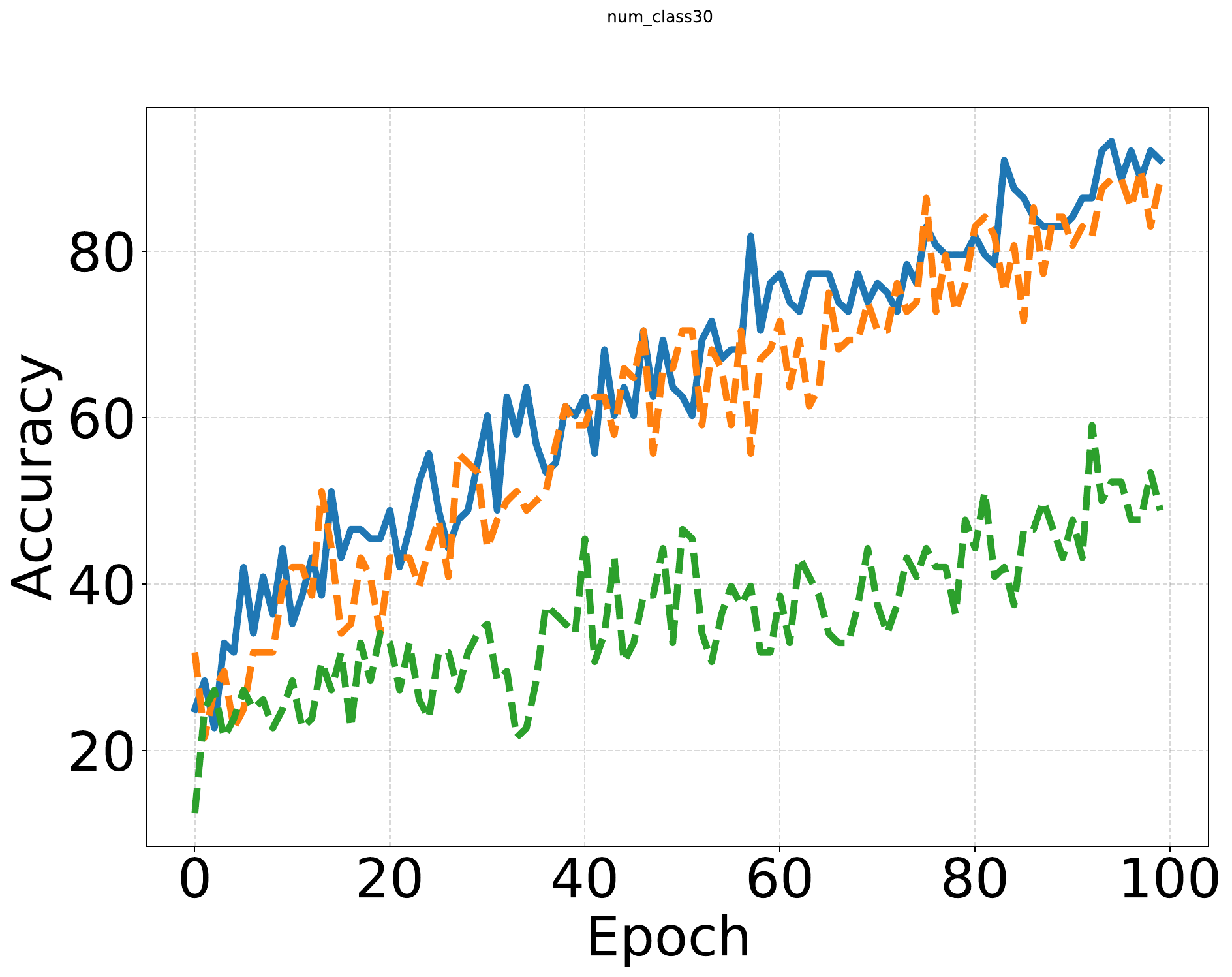}} 
    \subfigure[VGG, 40-shot]{
    \label{fig:VGG40-shot}
    \includegraphics[width=0.26\textwidth,trim=1 1 1 60,clip]{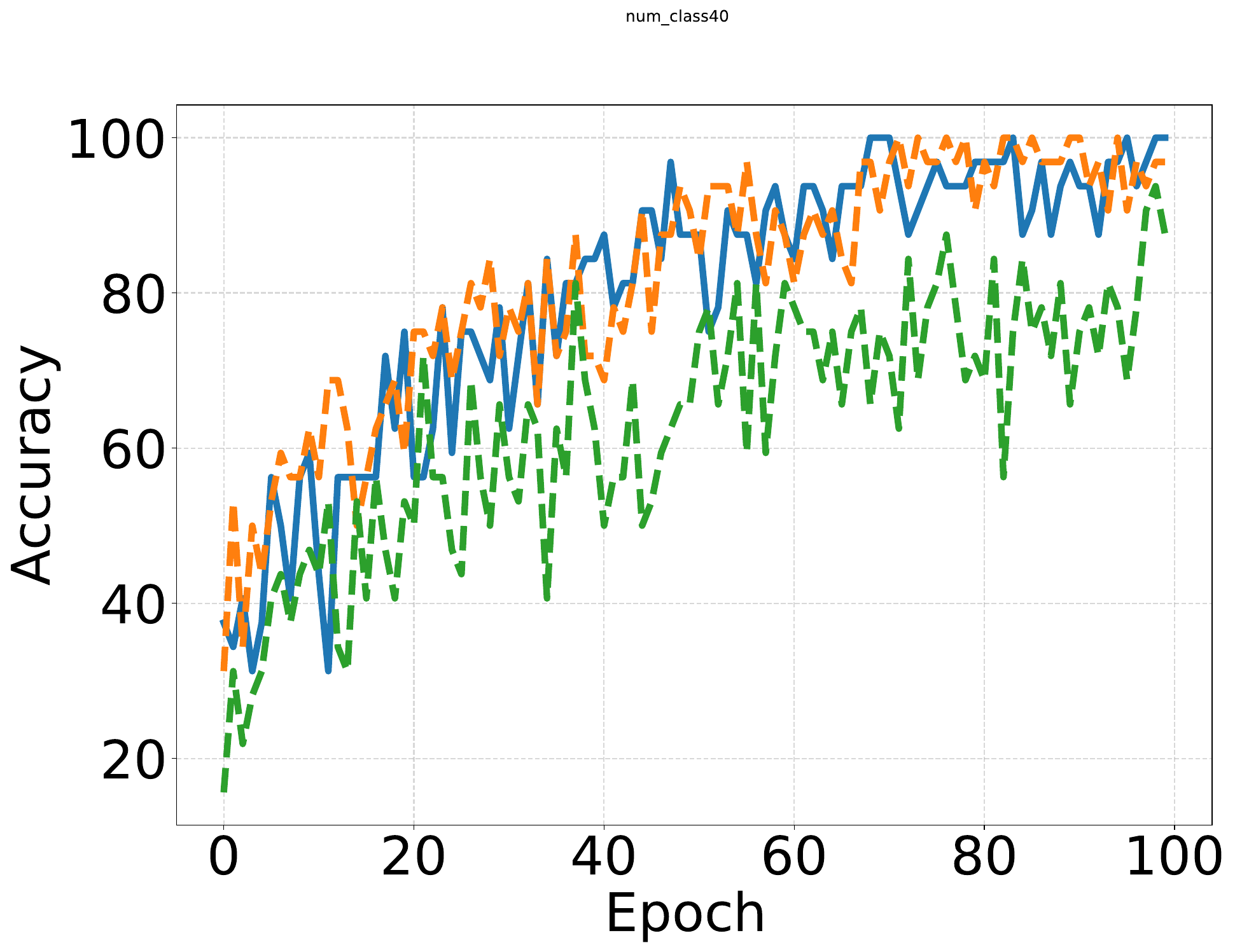}} 
    \subfigure[VGG, 50-shot]{
    \label{fig:VGG50-shot}
    \includegraphics[width=0.26\textwidth,trim=1 1 1 60,clip]{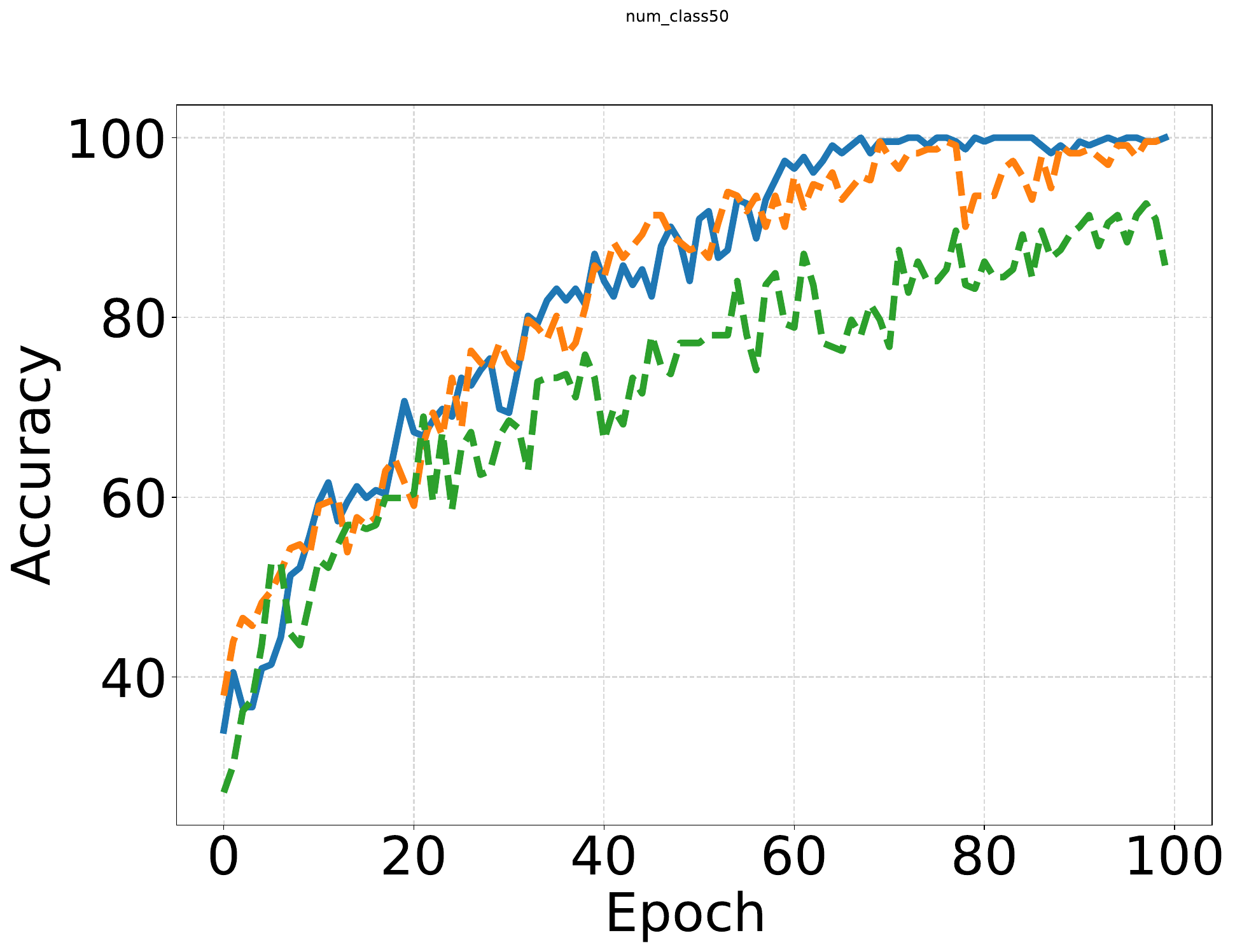}} 
    \subfigure[VGG, 60-shot]{
    \label{fig:VGG60-shot}
    \includegraphics[width=0.26\textwidth,trim=1 1 1 60,clip]{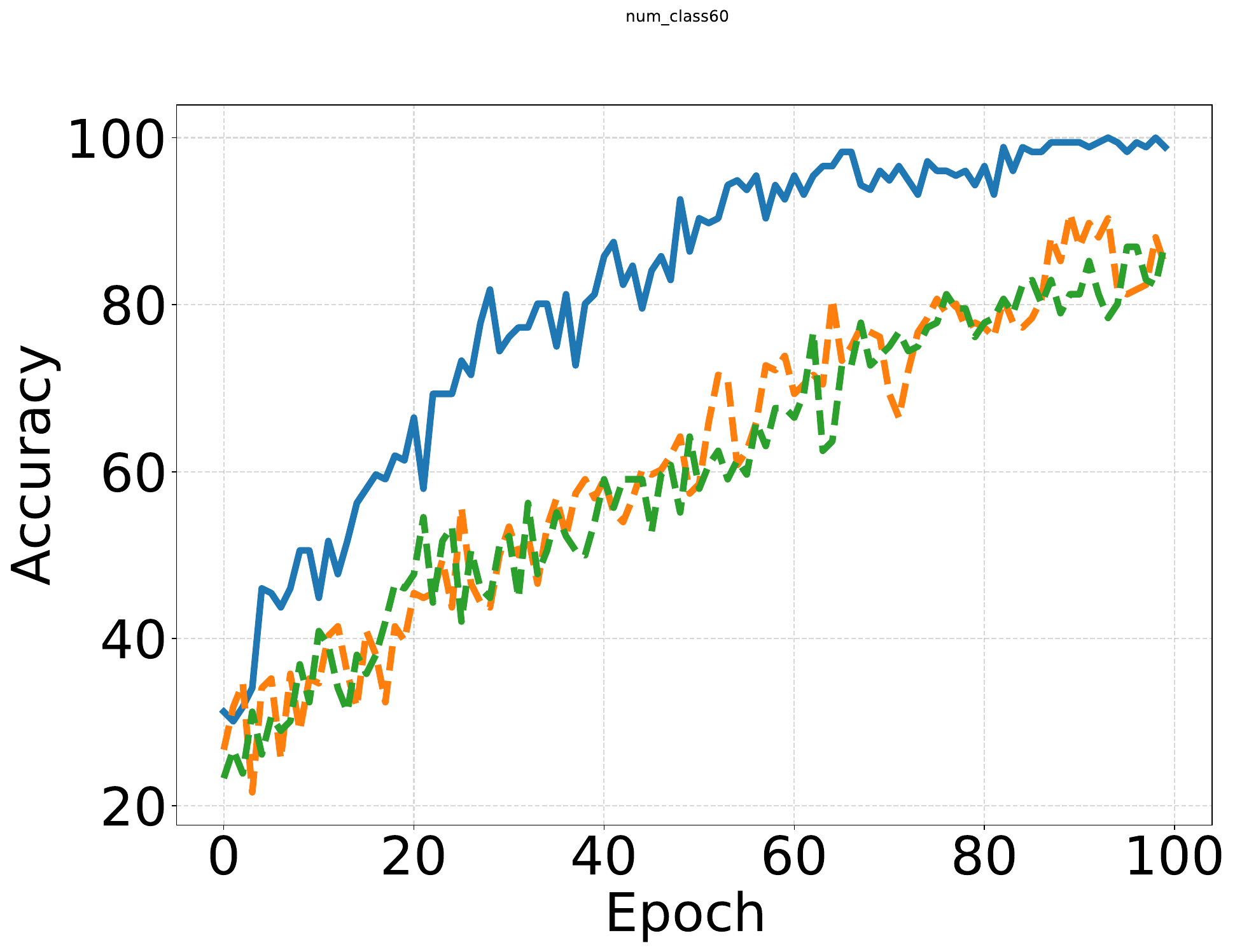}} 
    \subfigure[ResNet, 10-shot ]{
    \label{fig:ResNet10-shot}
    \includegraphics[width=0.26\textwidth,trim=1 1 1 60,clip]{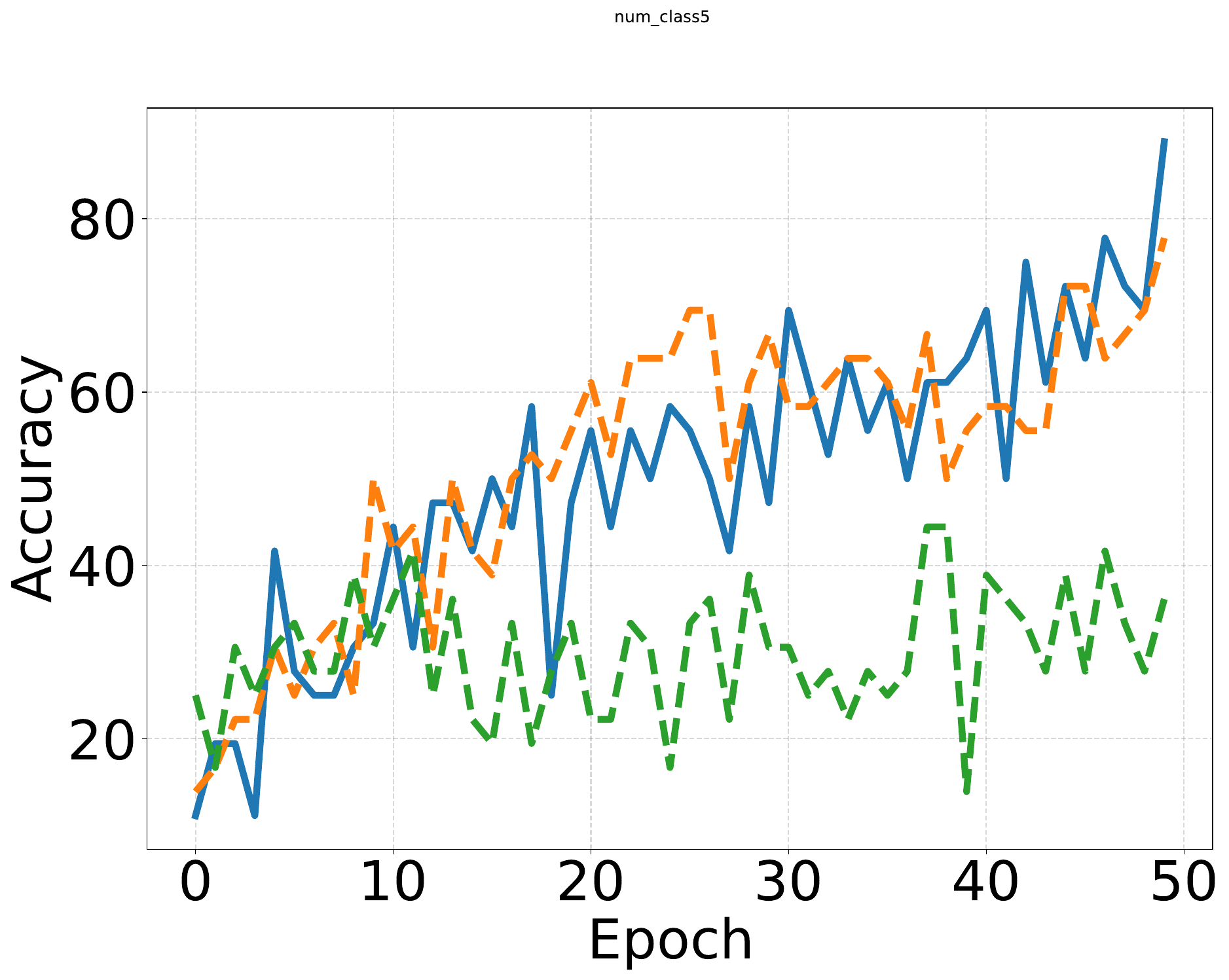}} 
    \subfigure[ResNet, 20-shot ]{
    \label{fig:ResNet20-shot}
    \includegraphics[width=0.26\textwidth,trim=1 1 1 60,clip]{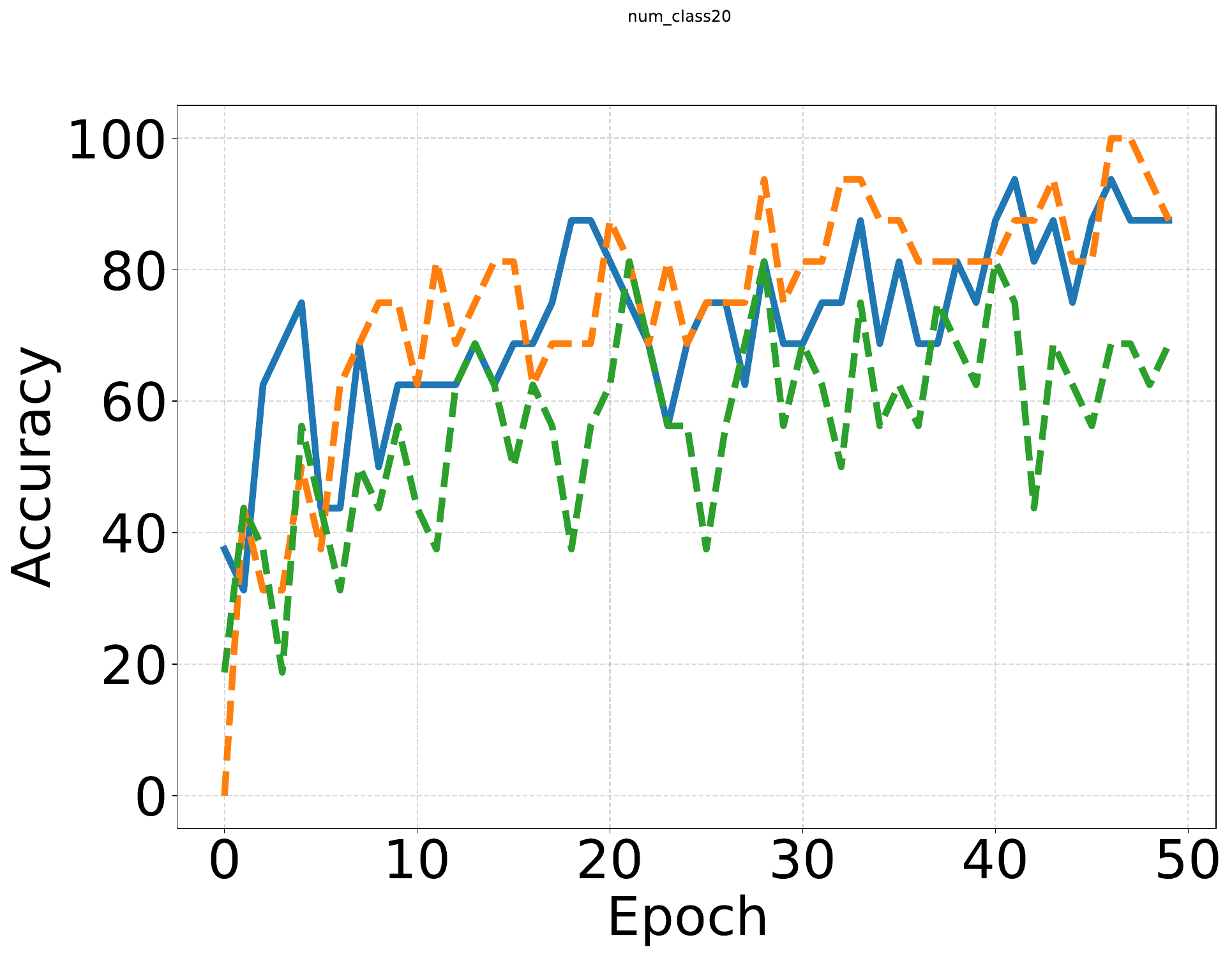}} 
    \subfigure[ResNet, 30-shot]{
    \label{fig:ResNet30-shot}
    \includegraphics[width=0.26\textwidth,trim=1 1 1 60,clip]{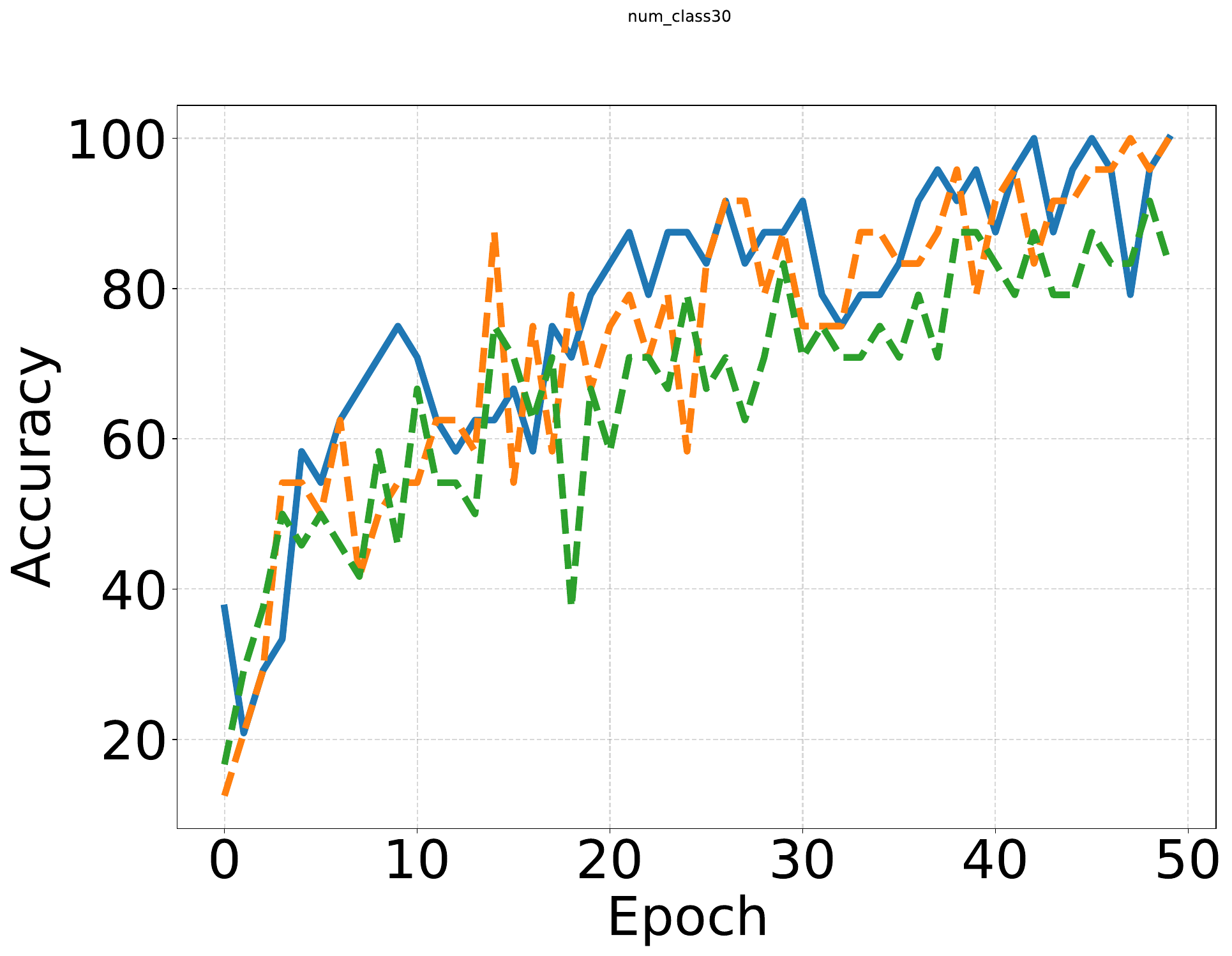}} 
    \subfigure[ResNet, 40-shot]{
    \label{fig:ResNet40-shot}
    \includegraphics[width=0.26\textwidth,trim=1 1 1 60,clip]{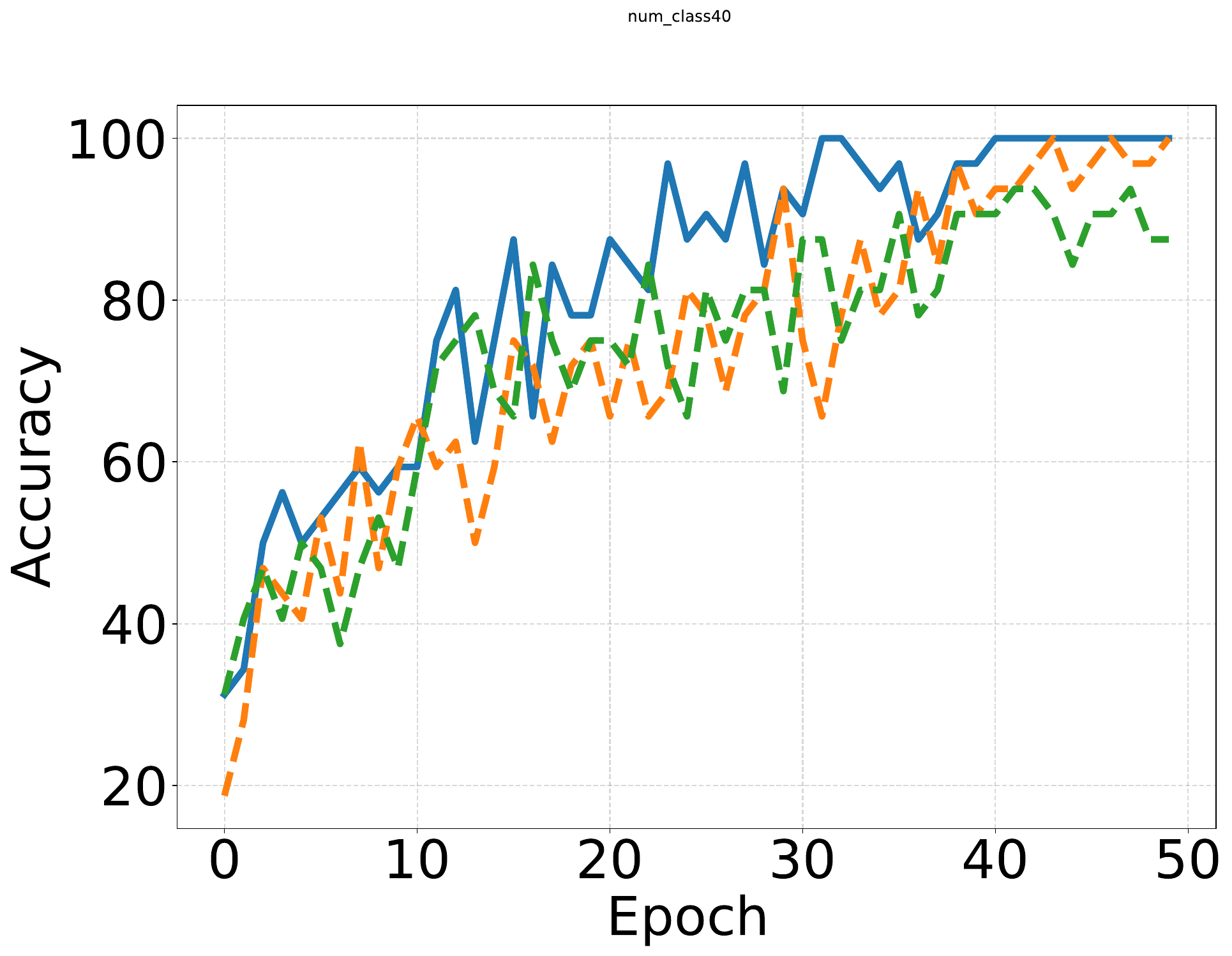}} 
    \subfigure[ResNet, 50-shot]{
    \label{fig:ResNet50-shot}
    \includegraphics[width=0.26\textwidth,trim=1 1 1 60,clip]{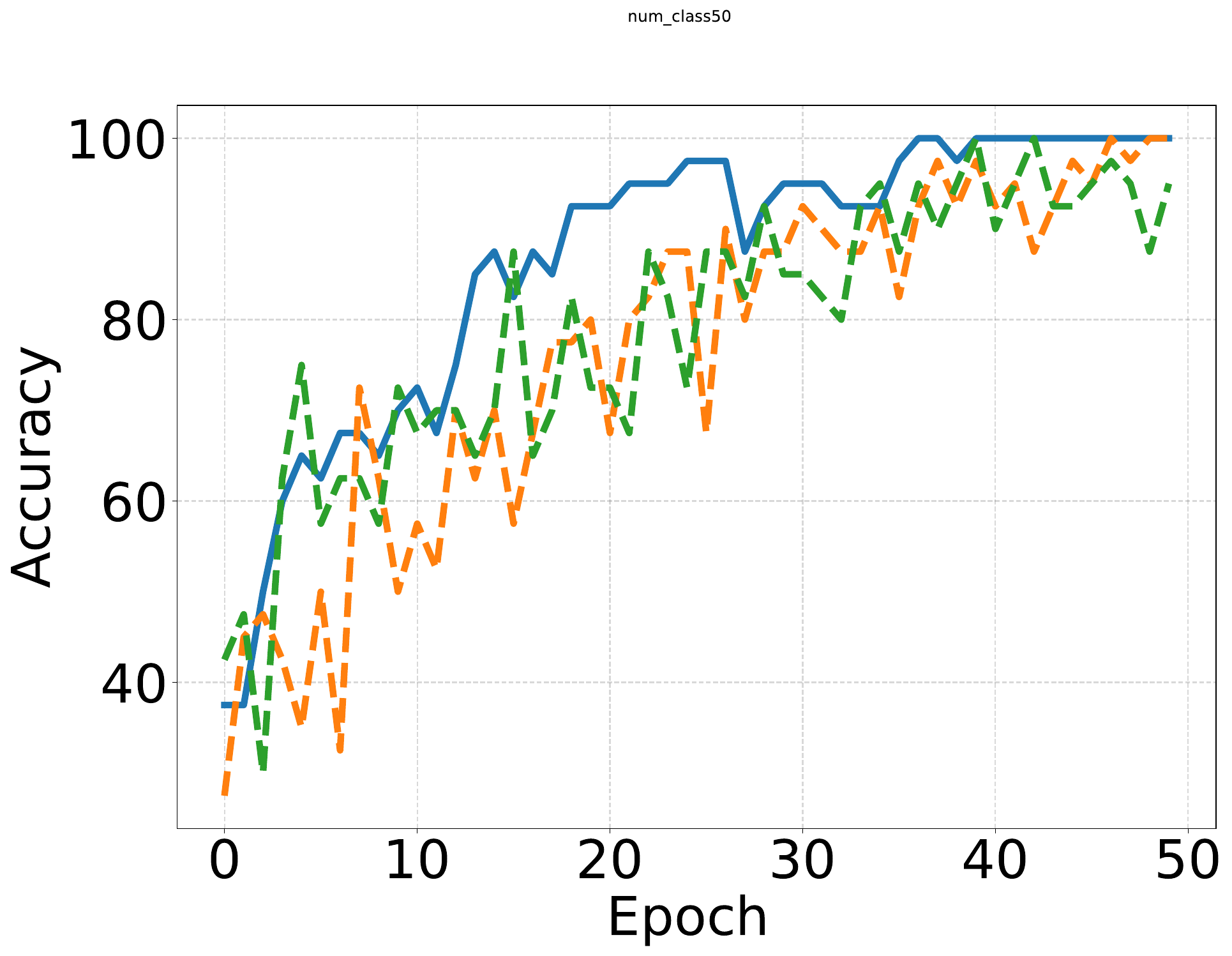}} 
    \subfigure[ResNet, 60-shot]{
    \label{fig:ResNet60-shot}
    \includegraphics[width=0.26\textwidth,trim=1 1 1 60,clip]{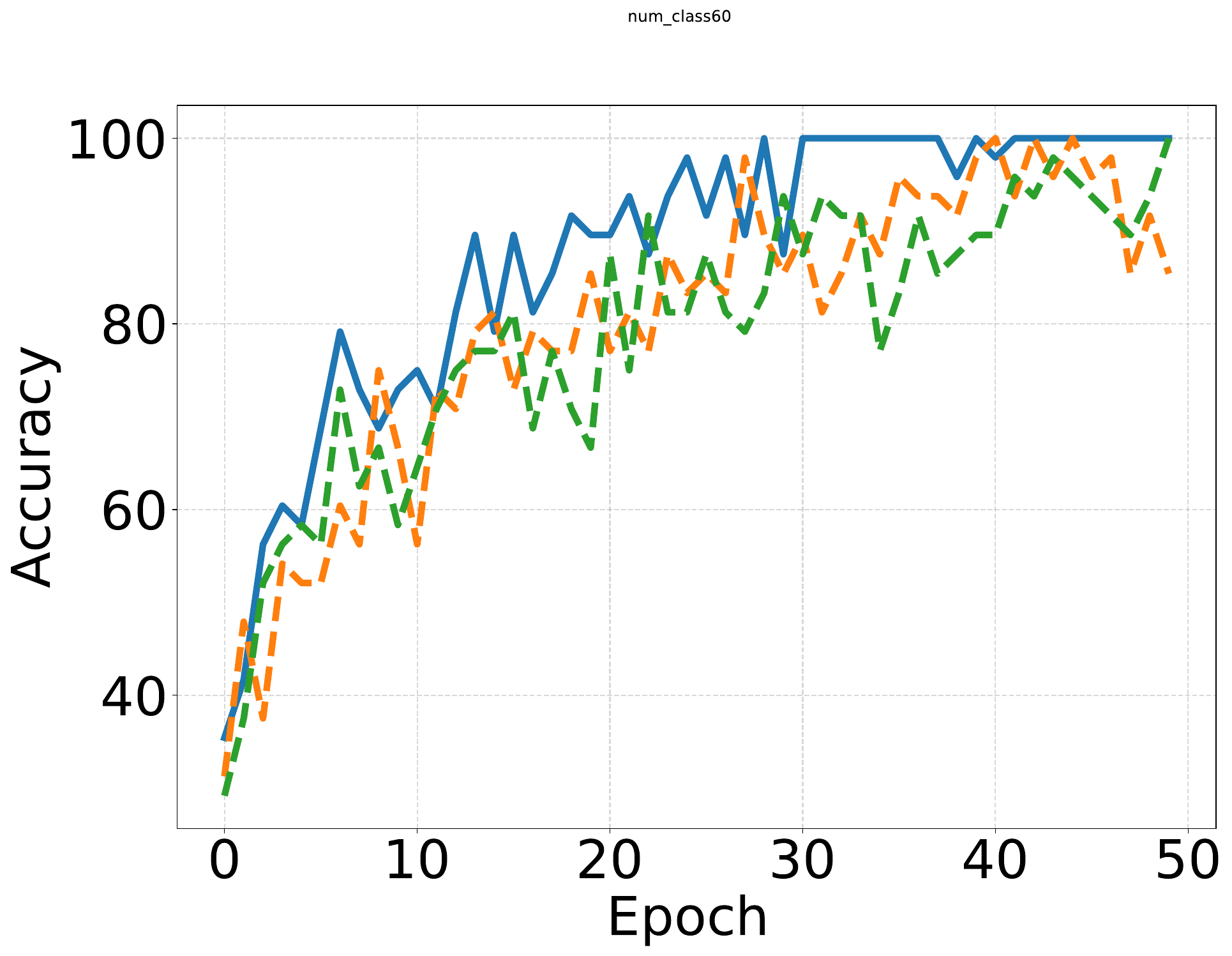}} 
  \caption{Training process of Auto-Learngene and other methods for various network architectures when using the different numbers of samples per class. Figure (a-r) reveal that our method is more efficient and converges faster. }
    \label{fig:R1}
    \vspace{-0.1in}
\end{figure*}

\section{Experiments}
\label{experiment}

\subsection{Experimental Setting}
\textbf{Datasets. } 1)\emph{ CIFAR100.} CIFAR100~\cite{krizhevsky2009learning} consists of 100 object classes and 60,000 images. We use 64 classes to train the ancestry model, 16 classes to automatically extract the learngene layers, and the remaining 20 classes to train the descendant model. Note that these classes do not overlap in three cases. 2)\emph{ ImageNet100. } ImageNet100~\cite{wang2021learngene} has 100 object classes and 60,000 color images of size 84×84. Similarly, ImageNet100 is also divided into three parts: 64, 16, and 20 for training the ancestry model, extracting the learngene layers, and training the descendant model, respectively. 3)\emph{ ImageNet. } ImageNet is a 1,000-class dataset
from ILSVRC2012~\cite{5206848}, providing 1.2 million images for training and 50,000 images for validation. We use it to establish the self-supervised learning~\cite{CaronTMJMBJ21}  for the ancestry model of the ViT backbone. 

\textbf{ Network architectures} 1)\emph{ Ancestry model.} We train the multiple big ancestry models, such as ViT~\cite{dosovitskiy2021an}, VGG16~\cite{DBLP:journals/corr/SimonyanZ14a} and ResNet18~\cite{he2016deep}.  
2)\emph{ Descendant model. } For ViT, our algorithm inherits the lower six encoder layers as the learngene layers and stacks them to a classification head of randomly initialized parameters. As a result, the descendant model is only half the size of the ancestry model. Considering the lightweight nature of the descendant model, we create VGG8 and ResNet12 as the descendant model. Our method automatically inherits the lower layers (\ie the convolutional layers with 64 output channels) and deeper layers (\ie the convolutional layers with 512 output channels), and then stacks them to some intermediate layers (\ie the convolutional layers with 128 and 256 output channels) and the classification head.~\footnote{
See Supplementary A.3 for more details} 3) \emph{ Meta-network. } For all experiments, a single fully-connected layer is used to construct the meta-network for each pair (l,k). It takes the output features $\bm{Z}^l$ of the ancestry model as input and outputs the layer similarity score $\alpha^{l,k}$. Moreover, the activation function is RELU6~\cite{krizhevsky2010convolutional}, which is defined as $\min (\max (x, 0), 6)$, ensuring that the output $\alpha^{l,k}$ is non-negative and not excessively large.


\textbf{Hyperparameters }
For CNNs, we set the learning rate to 0.1 with MultiStepLR and batch size to 64 on Pytorch~\cite{paszke2019pytorch} when training the ancestry model. ViT-based ancestry model uses the same hyperparameters as~\cite{CaronTMJMBJ21}.
The learning rate of the meta-network is uniformly fixed as $10^{-4}$ with CosineAnnealingLR. Furthermore, we set the batch size to 16 on the descendant model. 

\subsection{Experimental Results for Verifying the Advantages of Learngene }
\label{experimental_analysis}   
In this section, we undertake experiments to corroborate the four key benefits of \textit{Learngene}: (\rmnum{1}) \textbf{Faster Convergence}: \textit{Learngene} enables descendant models to significantly accelerate the training process. (\rmnum{2}) \textbf{Less Sensitivity to Hyperparameters}: \textit{Learngene} makes model performance less sensitive to the choice of  hyperparameters, \eg learning rate and weight decay. (\rmnum{3}) \textbf{Better Performance}: By efficiently initializing descendant models, \textit{Learngene} enables them to achieve better results in the specific tasks. (\rmnum{4}) \textbf{Fewer samples in downstream tasks}: \textit{Learngene} is inherited to descendant models to make them adapt to the downstream tasks with fewer samples. For convenience, our \textit{Learngene} algorithm is marked as \emph{Auto-Learngene}. Besides, two important comparison methods are described below: \emph{From-Scratch} directly randomly initializes descendant models and  \emph{Heur-Learngene}~\cite{wang2021learngene} employs a gradient-based heuristic approach to extract the learngene layers in the conference version.


\subsubsection{Faster Convergence.}
\label{exp:Convergence}
\textit{Learngene} is effective at providing a strong starting point for the descendant model, allowing it to find a satisfactory solution to the problem at hand more quickly, \ie significantly improving convergence speed. In our experiments, we report the results for various network architectures when using the different numbers of samples per class in Figure \ref{fig:R1} (a-r). We report the results by averaging five descendant models trained on five tasks. These include VGG on CIFAR100, as well as ResNet and ViT on ImageNet100. 

For the ViT, (a-f) show that the descendant models consistently exhibit quick convergence on the diverse samples, requiring only 20 epochs, whereas From-Scratch and Heur-Learngene require almost twice as many epochs to converge. Additionally, 
learning curves in (j-l, p-r) show that the CNN-based descendant models in the Auto-Learngene are the fastest to converge on  the target tasks, requiring only 60 epochs for the VGG and 30 epochs for the ResNet. In contrast, From-Scratch requires more than 100 or 50 epochs to converge for the VGG or the ResNet, respectively. In the extreme case, as shown by (g, m), From-Scratch struggles to converge. This is due to the lack of an effective initialization for From-Scratch, making it difficult to quickly fit the limited number of samples.



\begin{table}
  \caption{Performance of different initialization methods with different architectures.}
  \label{table:initialize}
  \centering
\begin{tabular}{ccccc}
\hline
\toprule[1pt]
Dataset     & Method              & ViT  & VGG   & ResNet   \\ \hline
            & From-Scratch         & 68.00     & 81.88 & 80.92   \\
CIFAR100    & Feature Distillation~\cite{zhang2021self} & 72.60 & 82.00 & 81.14   \\
            & Heur-Learngene~\cite{wang2021learngene}   & 69.36    & 82.78 & 80.92   \\
            &  Auto-Learngene     & \textbf{86.40}  & \textbf{84.04} & \textbf{84.56}   \\ \hline
            & From-Scratch            & 67.68   & 75.70 & 82.93  \\
ImageNet100 & Feature Distillation~\cite{zhang2021self} & 71.44 & 76.25 & 83.64   \\
            & Heur-Learngene~\cite{wang2021learngene}    & 68.56   & 76.80 & 82.45   \\
            &  Auto-Learngene     & \textbf{87.08}  & \textbf{77.60} & \textbf{85.45}   \\ 
\bottomrule[1pt]
\hline
\end{tabular}
\end{table}

\begin{figure*}[htbp]
\setlength{\abovecaptionskip}{0pt}
\setlength{\belowcaptionskip}{10pt}
\begin{center}
\centerline{\includegraphics[width=1.0\textwidth,trim=0 450 0 400,clip]{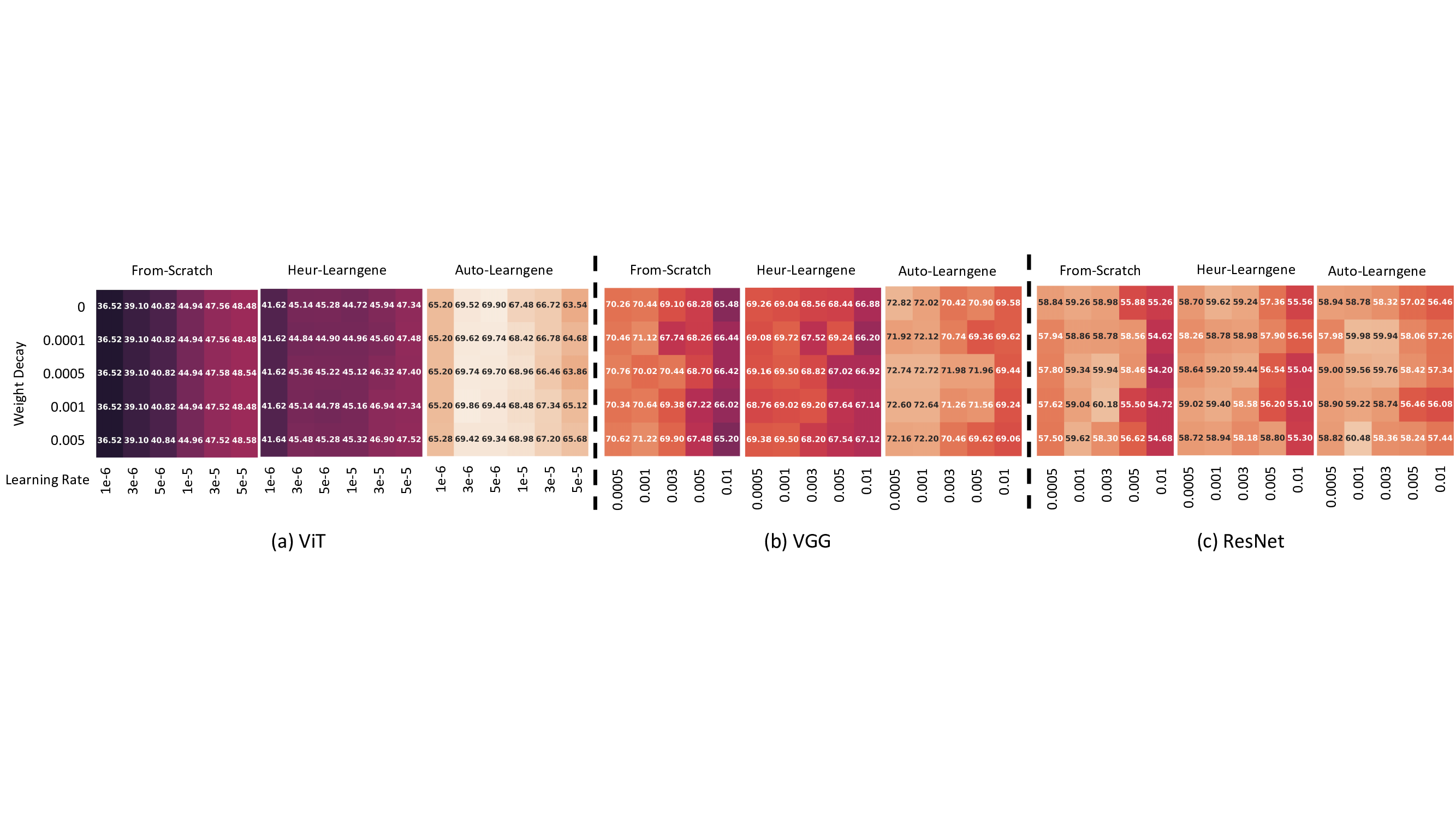}}
\caption{Learngene makes performance less sensitive to hyperparameters. We train the descendant models on various hyperparameter choices for ViT and CNNs. }
\label{fig:sensitive} 
\end{center}
\vskip -0.3in
\end{figure*}


\subsubsection{Less Sensitivity to Hyperparameters.}
\textit{Learngene} offers a more stable and consistent model performance, particularly in challenging scenarios where it is hard to determine the optimal hyperparameter values. In our experiments, we observe that \textit{Learngene} makes model performance less sensitive to hyperparameters of learning rate and weight decay, reducing the need for detailed hyperparameter tuning. To exhaustively verify that \textit{Learngene} is indeed insensitive to hyperparameters, we train descendant models with different network architectures: VGG for Cifar100, ResNet, and ViT for ImageNet100. Each model is trained on a five-category task with 50 samples per class for a total of 50 epochs. 

As shown in Figure~\ref{fig:sensitive}, we find that both the Auto-Learngene and Heur-Learngene exhibit fewer variations in performance over the different hyperparameters compared to From-Scratch. In the case of ViT, 
the accuracy of From-Scratch fluctuates by more than 11\%  when the learning rate changes from 1e-6 to 5e-5, but the accuracy of Auto-Learngene changes by less than 7\%. These results are consistent with previous findings~\cite{zhu2021gradinit, huang2020improving} which suggest that a model with good initialization does not require complex parameter tuning. Similarly, in (b-c), this is particularly evident for the Auto-Learngene, where the performance of From-Scratch 
on VGG degrades to 65.2\% and on ResNet to 54.68\% with a learning rate of 0.01 and weight decay of 0.005, while the accuracy of Auto-Learngene remains above 69\% on VGG and 57\% on ResNet.

\subsubsection{Better Performance.}
\label{exp:other_initialization}
We compare \textit{Learngene} with other initialization methods, \eg From-Scratch, Heur-Learngene~\cite{wang2021learngene}, Feature Distillation~\cite{zhang2021self}. 
For a fair comparison, Feature Distillation implements the transfer of outputs from the selected layers in the source model to the target model, and these layers correspond precisely to the locations of the learngene layers as identified by Auto-Learngene. The training settings of the source model for Feature Distillation and Heur-Learngene are consistent with those of the ancestry model for Auto-Learngene, \eg a VGG model trained on CIFAR100, a ResNet model trained on ImageNet100, or a ViT model trained on ImageNet. Moreover, we set the size of the target model for other initialization approaches to be identical to that of the descendant models for Auto-Learngene.

In Table~\ref{table:initialize}, we report the results by averaging  the performance of five descendant models on CIFAR100 and ImageNet100, respectively. Each model is trained on a five-category task with 500 samples per class for a total of 50 epochs. One can observe that Auto-Learngene compares favorably with all these baseline algorithms. In particular, for ViT, our algorithm significantly improves accuracy when compared to other initialization methods: Auto-Learngene improves over 13\% on CIFAR100 and 15\% on ImageNet100, respectively. Therefore, we provide evidence that the same learngene can be inherited into the descendant models with different architectures, even when the data of the source and target domains do not share the same distribution. For VGG, the best test accuracies of other initialization methods are 82.78\% and 76.80\% on CIFAR100 and ImageNet100, respectively, while Auto-Learngene  achieves 84.04\% and 77.60\%, respectively. Similar results are observed in the ResNet architecture. Several recent studies~\cite{park2022how, liu2021efficient, Gani2022HowTT} have shown that a proper initialization significantly affects the performance of ViT trained in small data regimes. Therefore, this finding explains why the advantages of \textit{Learngene} are more pronounced in ViT as compared to those in CNNs.

\begin{figure*}[htbp]
\begin{center}
\centerline{\includegraphics[width=1.0 \textwidth,trim=0 140 0 110,clip]{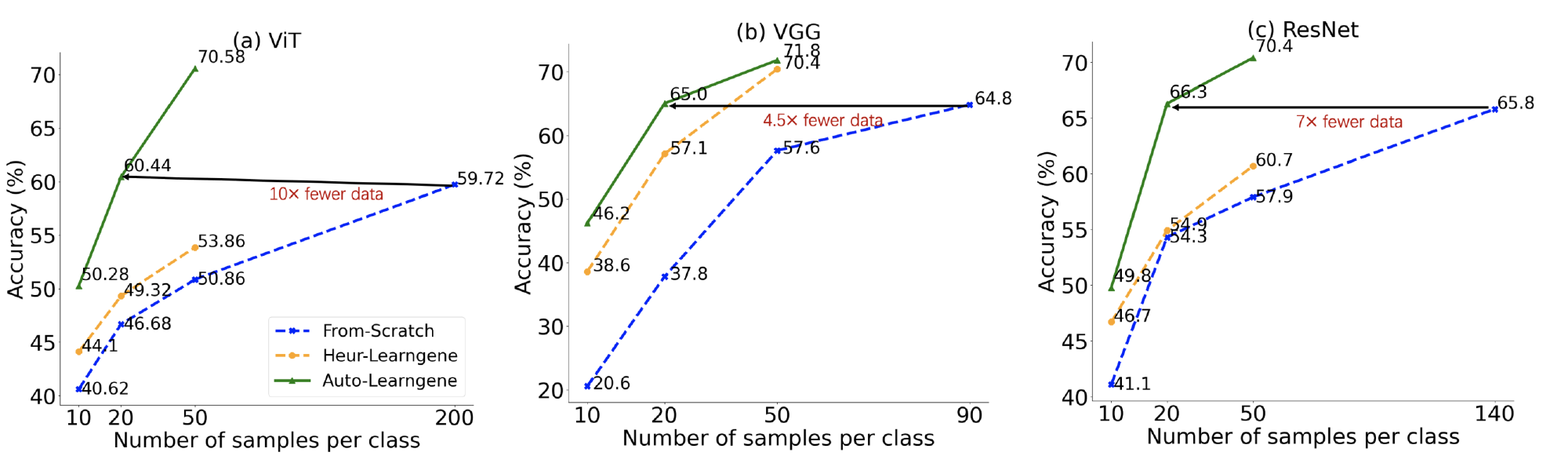}}
\caption{\emph{Performance of Auto-Learngene and other methods for various network architectures when using diverse amounts of data.}}
\label{fig:fewer_samples}
\end{center}
\vspace{-0.3in} 
\end{figure*}

\begin{table*}
  \caption{Multi-class classification accuracies on CIFAR100 and ImageNet100.}
  \label{accuracy-table}
  \centering
  \begin{small}
\begin{tabular}{ccccc} 
\hline
\toprule[1pt]
                 & \multicolumn{2}{c}{CIFAR100, 5way, VGG} & \multicolumn{2}{c}{ImageNet100, 5way, ResNet} \\
Method           & 10-shot            & 20-shot            & 10-shot                & 20-shot               \\ \hline
MatchingNet~\cite{vinyals2016matching}      & 56.71$\pm$4.48     & 60.98$\pm$3.09     &  53.14$\pm$2.78         & 60.48$\pm$2.66       \\
MAML~\cite{finn2017model}            & 51.70$\pm$1.75     & 62.36$\pm$2.88     & 48.73$\pm$3.10         & 57.10$\pm$2.42        \\
ProtoNet~\cite{snell2017prototypical}        & 56.40$\pm$2.88     & 61.02$\pm$3.18     & 52.91$\pm$1.57         & 60.13$\pm$1.73        \\
Reptile~\cite{nichol2018first} & 50.13$\pm$3.73     & 60.00$\pm$3.31     & 46.52$\pm$2.63         & 54.70$\pm$1.96        \\
Baseline++~\cite{Chen2019ACL}      & 52.67$\pm$4.02     & 61.20$\pm$3.29     & 53.24$\pm$2.75         &  62.43$\pm$1.63        \\
DeepEMD~\cite{zhang2020deepemd}         & 58.03$\pm$1.85     & 63.69$\pm$2.05     & 54.93$\pm$1.78         &  63.41$\pm$1.47       \\
ProtoMAML~\cite{Triantafillou2020Meta-Dataset}      & 53.11$\pm$1.25     & 59.57$\pm$1.34     & 52.74$\pm$2.52         & 62.23$\pm$1.34        \\
First-order MTL~\cite{wang2021bridging}       & 57.12$\pm$3.29     &  63.30$\pm$2.24    &  52.20$\pm$2.60        & 63.10$\pm$2.11 \\
BOIL~\cite{oh2021boil} &  59.14$\pm$2.51    &  65.01$\pm$1.35   & 55.40$\pm$2.43        &  64.85$\pm$1.28
\\ \hline
Heur-Learngene~\cite{wang2021learngene}   & 58.86$\pm$3.12     & 64.24$\pm$1.89     & 54.00$\pm$3.33         & 63.40$\pm$1.19        \\
\rowcolor{gray!40} Auto-Learngene   & \textbf{61.80}$\pm$2.62     & \textbf{68.48}$\pm$2.75     & \textbf{55.76}$\pm$4.68         & \textbf{66.04}$\pm$2.09        \\ 
\bottomrule[1pt]
\hline
\end{tabular}
\end{small}
\end{table*}

\subsubsection{Fewer samples in downstream tasks.}
The last advantage of \textit{Learngene} is that different descendant models, when initialized by it, can quickly adapt to different tasks using a few training samples. Our experiments demonstrate that the descendant models initialized by \textit{Learngene} significantly reduce the number of required training samples. We train descendant models using different network architectures (\eg ViT and ResNet for ImageNet100, VGG for Cifar100) with varying numbers of samples. 

As shown in Figure~\ref{fig:fewer_samples}, the difference between \textit{Learngene} and From-Scratch is even more pronounced for the ViT on ImageNet100: Using 10$\times$ fewer data, our Auto-Learngene achieves 60.44\% accuracy, which is on-pair with From-Scratch. 
Moreover, the descendant model only needs 20 samples per class to achieve 65\% performance for the VGG on CIFAR100. By contrast, From-Scratch overfits under such condition, and the gap is obvious, where it attains at least 27\% lower average accuracy. More importantly, Auto-Learngene is able to achieve 65\% accuracy with only 2/9 of the data that From-Scratch used. Similar results are observed in the ResNet architecture. As previously noted in Section~\ref{exp:Convergence}, Auto-Learngene converges rapidly, indicating that it provides an effective initialization. 


In addition, we compare \textit{Learngene} to other few-shot learning methods. We implement our algorithm and compare it against the few-shot learning algorithm on CIFAR100 and ImageNet100, averaging the accuracy over 100 tasks. We set backbones of the same scale as the descendant models for ten baseline methods~\cite{vinyals2016matching, finn2017model, snell2017prototypical, nichol2018first, Chen2019ACL, zhang2020deepemd, Triantafillou2020Meta-Dataset, wang2021bridging, oh2021boil, wang2021learngene}. Since other baseline methods reuse the entire model, to ensure fair comparisons, we apply knowledge transformation~\cite{heo2019comprehensive
} to reuse as much knowledge as possible from the ancestor model to the randomly initialized layers in the descendant models. As shown in Table~\ref{accuracy-table}, Auto-Learngene achieves the best performance compared to \textit{non-Learngene} methods.  For example, on the 20-shot tasks, the best \textit{non-Learngene} BOIL performs 65.01\% on CIFAR100 and 64.85\% on ImageNet100, while Auto-Learngene achieves 68.48\% and 66.04\% respectively. The reason for this is that \textit{Learngene} preserves significance, allowing descendant models to exhibit enhanced generalization on downstream tasks. In contrast, other \textit{non-Learngene} methods that directly reuse the entire model may result in negative transfer~\cite{wang2019characterizing} on downstream tasks. Besides, Auto-Learngene consistently improves upon Heur-Learngene, \eg achieving a 2.94\% gain on 10-shot tasks and a 4.24\% gain on 20-shot tasks on CIFAR100, and a 1.76\% gain on 10-shot tasks and a 2.64\% gain on 20-shot tasks on ImageNet100.

\subsection{Evolution Process of Ancestry Model}
\label{exp:evolution}
From a biological perspective, the ancestry with longer evolutionary history passes down the gene to their descendants, who have a stronger ability to quickly adapt to new environments. Likewise, during the accumulating process, the ancestry model can employ a training setting akin to lifelong learning~\cite{Parisi2018ContinualLL} instead of pretraining. 
As the number of tasks trained by the ancestry model increases, the significance of tasks becomes more concentrated in the learngene. Consequently, the learngene extracted from the ancestry model of later tasks is inherited to different descendant models, which will perform better overall on the downstream tasks.

Inspired by this evolutionary perspective, we design the following experiments on different backbones: 1) In the ViT case, we randomly sample the tasks from ImageNet, each containing 50 classes, and sequentially train the ancestry model on a total of 25 tasks. After training each task for 300 epochs, we inherit the learngene to five descendant models and then average their performance on CIFAR100. 2) In the CNN case, taking VGG as an example, we randomly sample tasks from the 64 classes of CIFAR100, each task containing 5 classes, and train the ancestry model on 25 sequential tasks. After training each task for 100 epochs, the learngene is inherited to five descendant models and then we average their results. In both cases mentioned above, each descendant model is trained on a five-class task with 500 samples per class, amounting to 50 epochs of training.

\begin{figure}[htbp]
\vspace{-0.12in}
\centerline{\includegraphics[width=0.53\textwidth,trim=200 380 200 320,clip]{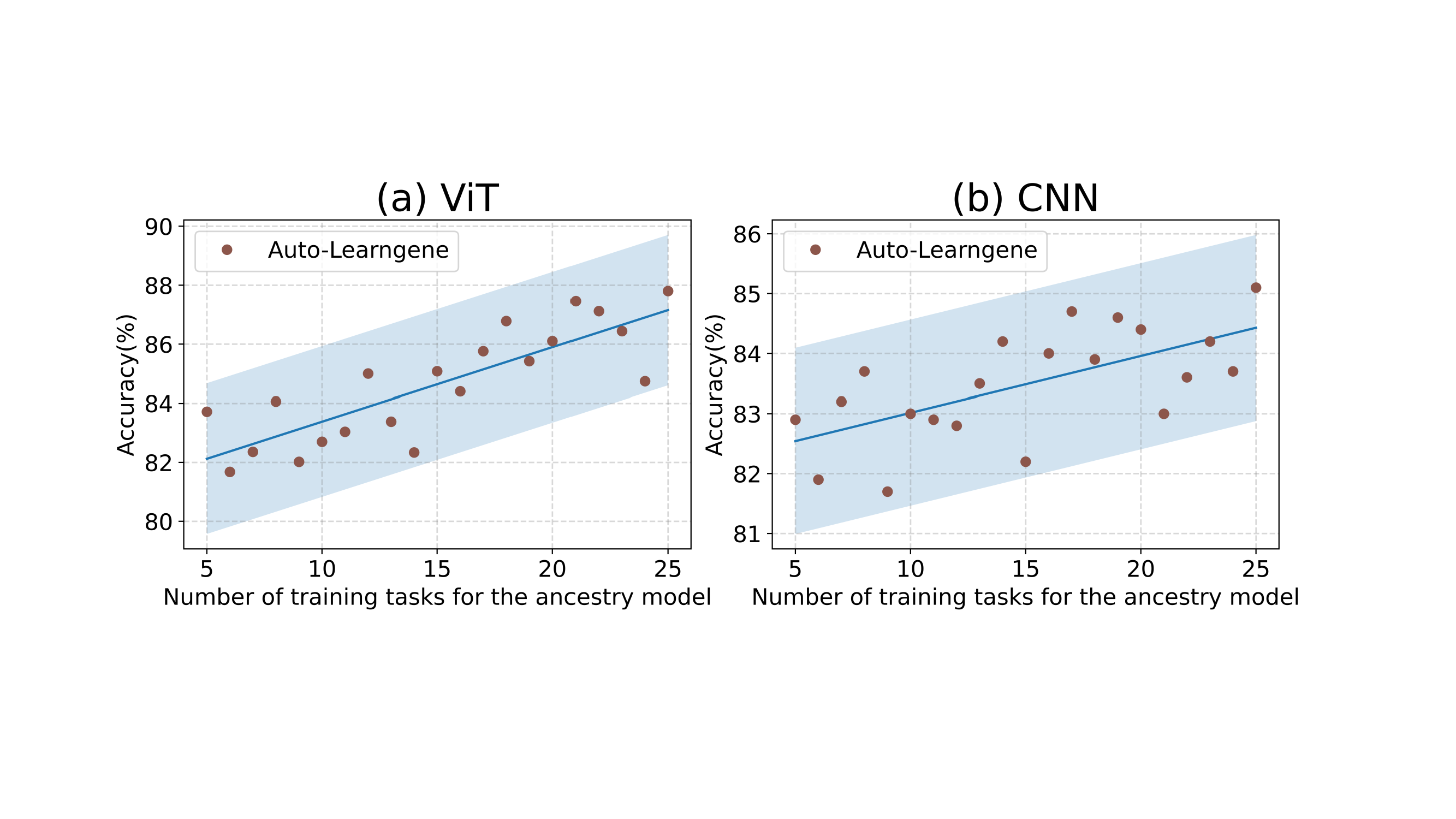}}
\caption{The performance of the descendant models shows a trend of continuous improvement when using different network architectures, as the ancestry model is trained on sequential tasks.} 
\label{fig:evolution}
\end{figure}

As illustrated in Figure~\ref{fig:evolution}, the performance of Auto-Learngene displays an upward trend as the number of tasks trained by the ancestry model increases. This can be attributed to \textit{Learngene}'s ability to preserve significant knowledge throughout the continuous training process of the ancestry model, akin to how a gene encapsulates the most stable part during evolution.

\begin{figure*}[htb]
\begin{center}
\centerline{\includegraphics[width=0.98 \textwidth,trim=150 500 250 450,clip]{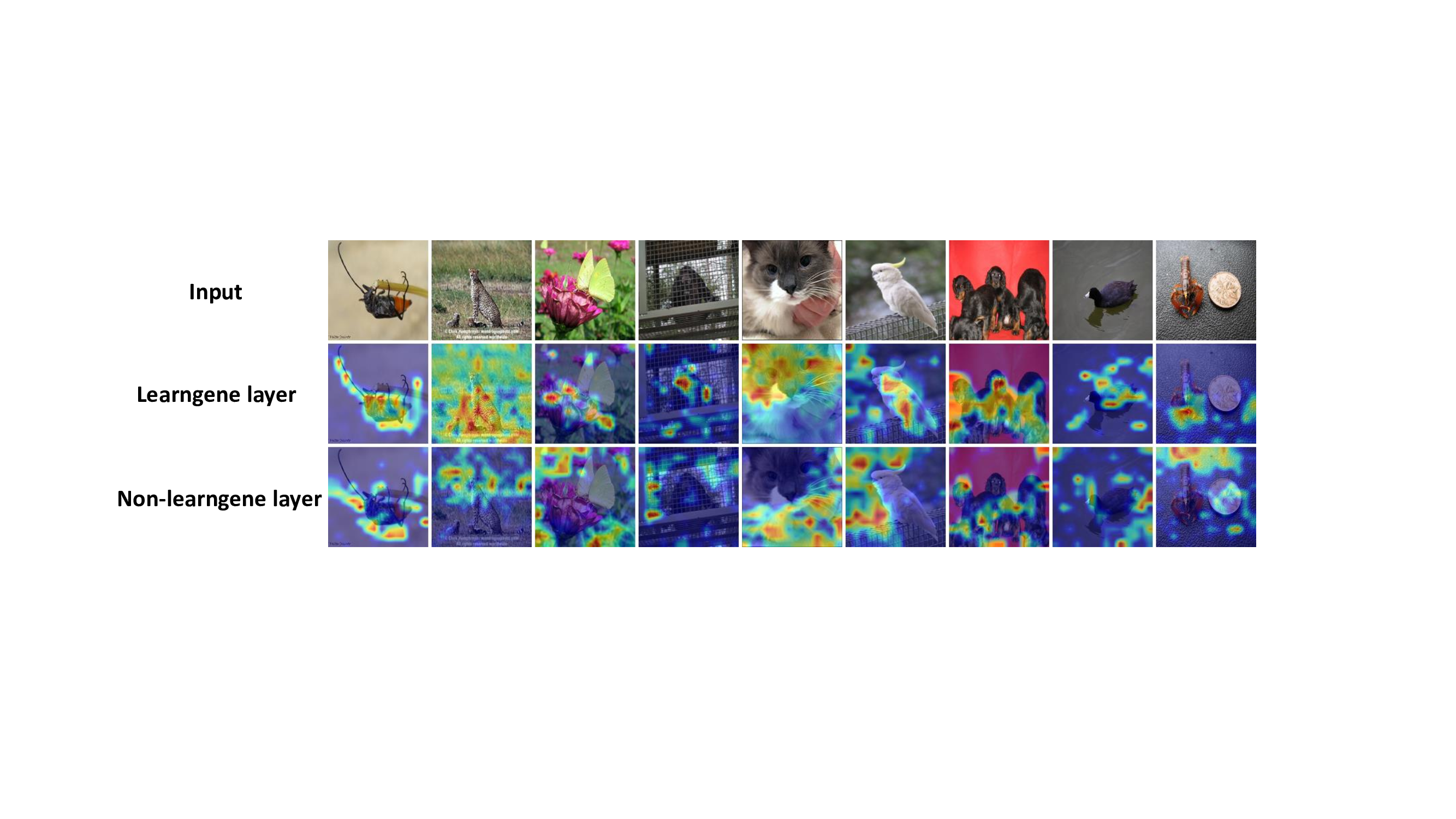}}
\caption{Visualization for the ViT-based ancestry model on ImageNet samples. The lower layers of an ancestry model are extracted as the learngene layers and the higher layers of the ancestry model are the non-learngene layers. This Figure illustrates that the learngene layers in ViT contain the local texture and semantic concept. }
\label{fig:visualization_vit}
\end{center}
\vspace{-0.1in} 
\end{figure*}

\begin{figure*}[htb]
\begin{center}
\centerline{\includegraphics[width=0.98 \textwidth,trim=150 300 200 250,clip]{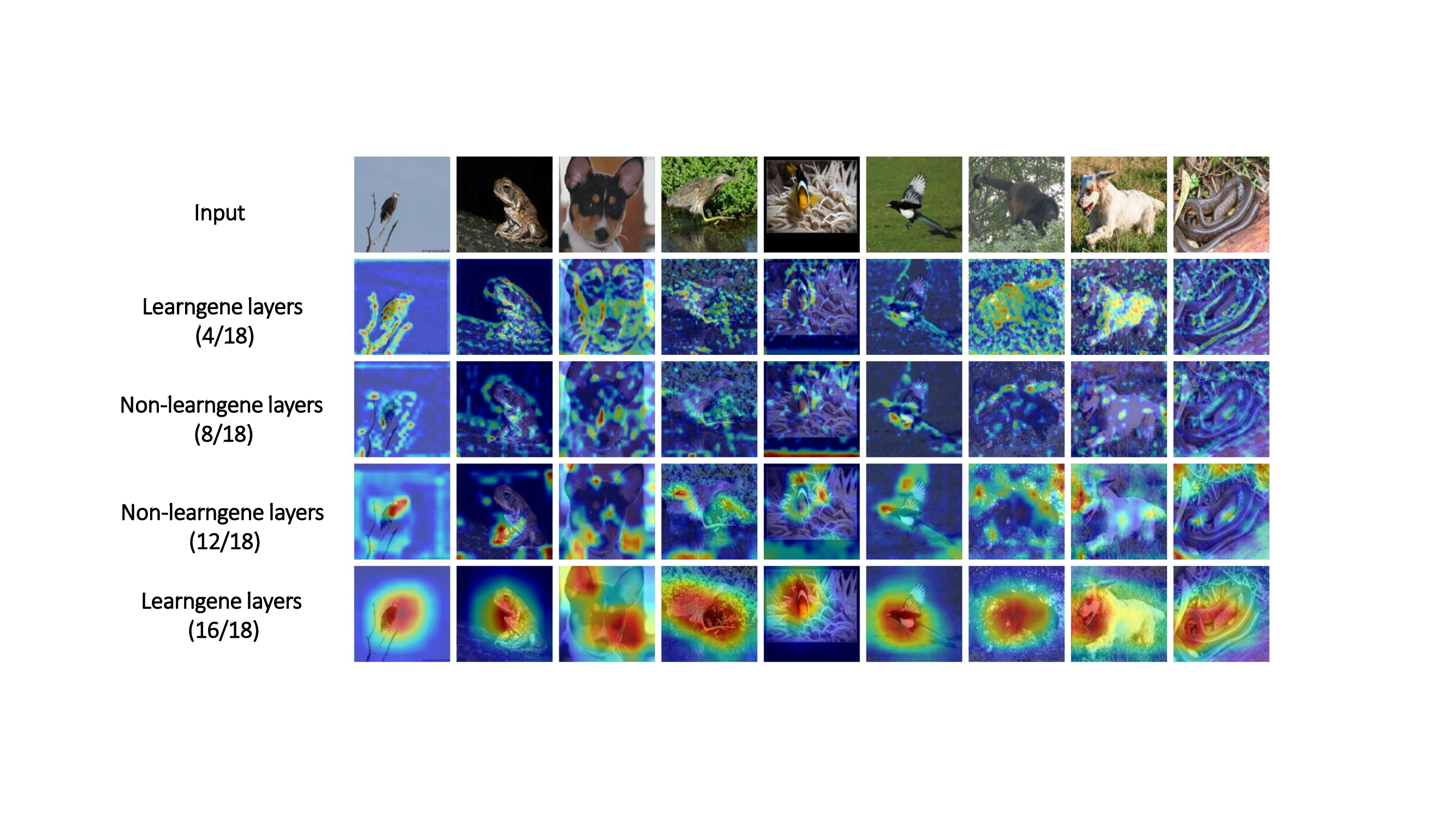}}
\caption{Visualization for the ResNet-based ancestry model on ImageNet samples. Note that $l$/18 means the $l$-th layer in ResNet18. The low and high layers of the ancestry model are chosen by the meta-network as the learngene layers. This figure shows that the learngene layers focus on the local texture and semantic concept, respectively.}
\label{fig:visualization} 
\end{center}
\vspace{-0.2in} 
\end{figure*}

\subsection{Qualitative Visualization}
\label{exp:why_work}
We use the visualization technique~\cite{selvaraju2017grad} to derive the output of different layers in the ancestry model and qualitatively visualize why \textit{Learngene} can preserve the significant knowledge. These visualizations can help us understand the role that the condensed learngene layers play in the overall performance of the model. 

For ViT, the meta-network is inclined to extract the lower layers of an ancestry model as the learngene layers, thereby rendering the higher layers of the ancestry model as the non-learngene layers. Some preceding studies~\cite{raghu2021do, park2022how} have shown that the lower layers in ViT encompass the local texture and semantic concept. Usually, the local texture and semantic concept are the significant knowledge~\cite{selvaraju2017grad, GilpinBYBSK18}. Figure~\ref{fig:visualization_vit} shows that the learngene layer is able to pay attention to both semantic concept and local texture, while the non-learngene layer is not able to accurately focus on these features. For example, in the first column of Figure~\ref{fig:visualization_vit}, the learngene layer is able to perceive the outline of the insect, while the non-learngene layer is not able to clearly identify it. Similarly, in the second column, the learngene layer is able to perceive both the semantic concept of a leopard and the local texture of the grassland environment where it is located, while the non-learngene layer is unable to do so. These findings suggest that the learngene layers of ViT can preserve significant knowledge.

In the case of CNNs, we take ResNet as an example. In this case, the low and high layers are chosen by the meta-network as the learngene layers, while the middle layers become the non-learngene layers. As illustrated in Figure~\ref{fig:visualization}, we observe that the learngene layers (4/18) pays attention to the local texture (\eg in the first column, the learngene layers (4/18) focuses on the silhouette of the bird and the surrounding environment of the tree) and the learngene layers (16/18) is more attuned to the semantic concept (\eg in the first column, the area of the bird in the picture from the learngene layers (16/18) is red). On the contrary, the area of interest in the image from the non-learngene layers is cluttered and less concentrated. This explains which significant knowledge our algorithm preserves in CNNs.

\section{Summary and discussions}
\label{summary}
Motivated by the fact that ancestral genetic information is inherited by newborns and aids in their rapid learning of new knowledge through just a few instances, we propose \textit{Learngene} to condense the ancestry model into a compact information piece and initialize the lightweight descendant models to quickly adapt to target tasks in low-data regimes. To this end, our paper lists some technical issues and presents the corresponding solutions. Experimental results show clear advantages of the \textit{Learngene}. 

Generally speaking, there are at least four scenarios where \textit{Learngene} could be helpful as follows:
\begin{itemize}
  \item [1)] 
  \textit{Learngene} initializes different descendant models on the target tasks of diverse domains from the training domain of the ancestry model, as illustrated in this paper. 
  
  \item [2)]
  In practical applications, clients can avoid resource-intensive tasks like data collection, model building, and infrastructure development. They simply need to obtain a learngene to initialize their models with a few samples. This not only protects the intellectual property of the model owners but also presents an appealing solution for privacy-conscious industries.
  
  \item [3)]
  \textit{Learngene} acts as a bridge from large models to small models, helping to generate lightweight models. Therefore, \textit{Learngene} provides a new way towards model compression~\cite{choudhary2020comprehensive}.
  
  \item [4)]
  As the ancestry model continues~\cite{delange2021continual} to be trained on an increasing number of tasks, the initialization capability of the learngene may progressively enhance.
\end{itemize}

The investigation of the \textit{Learngene} has only just begun and there are many open questions:
\begin{itemize}
  \item [1)] 
  Are there other forms of \textit{Learngene}? We have utilized various integral layers of diverse network architectures, \eg BatchNorm and Convolutional layers in CNNs, multi-head self-attention, LayerNorm, and Feed-Forward Network layers in ViT. In fact, any compact information pieces extracted from the ancestry model can be used as the learngene, \eg certain core neuron connections. 
  
  \item [2)]
  Are there other ways to implement learngene condensing? This paper leverages a meta-learning technique to automatically extract the learngene. Of course, when the form of \textit{Learngene} is defined, other dedicated methods can be used to extract it.
  
  \item [3)]
  How does the inherited learngene initialize the descendant model?  For our preliminary attempt, we have simply stacked learngene layers onto a few integral layers. Other methods may be designed to automatically extend the \textit{Learngene} to different descendant models for various downstream tasks.
  
\end{itemize}



%

\appendices
\section{Algorithm and implementation details}

\subsection{ Data Division in the Process of Learngene Condensing}

The CIFAR100 and ImageNet100 are divided into 64, 16 and 20 subsets  for training the ancestry model, extracting learngene layers, and training the descendant model, respectively. Moreover, the data used for selecting the learngene layers is divided into $1/6$ for meta-data $\widehat{\mathcal{D}}$ and $5/6$ for training data $\mathcal{D}$.


\subsection{ The total Amount of Computing}
As aforementioned in the full paper, Auto-Learngene saves computing sources in comparison with Heur-Learngene. We re-implemented Heur-Learngene on CIFAR100 using a V100 and it took about 13 hours and 37 minutes. In the same setting, Auto-Learngene runs about 8 hours and 45 minutes, which saves 1/3 of the computational resources.

\subsection{Model description}
\begin{figure*}[htb]
\begin{center}
\centerline{\includegraphics[width=1.0 \textwidth,trim=0 0 0 0,clip]{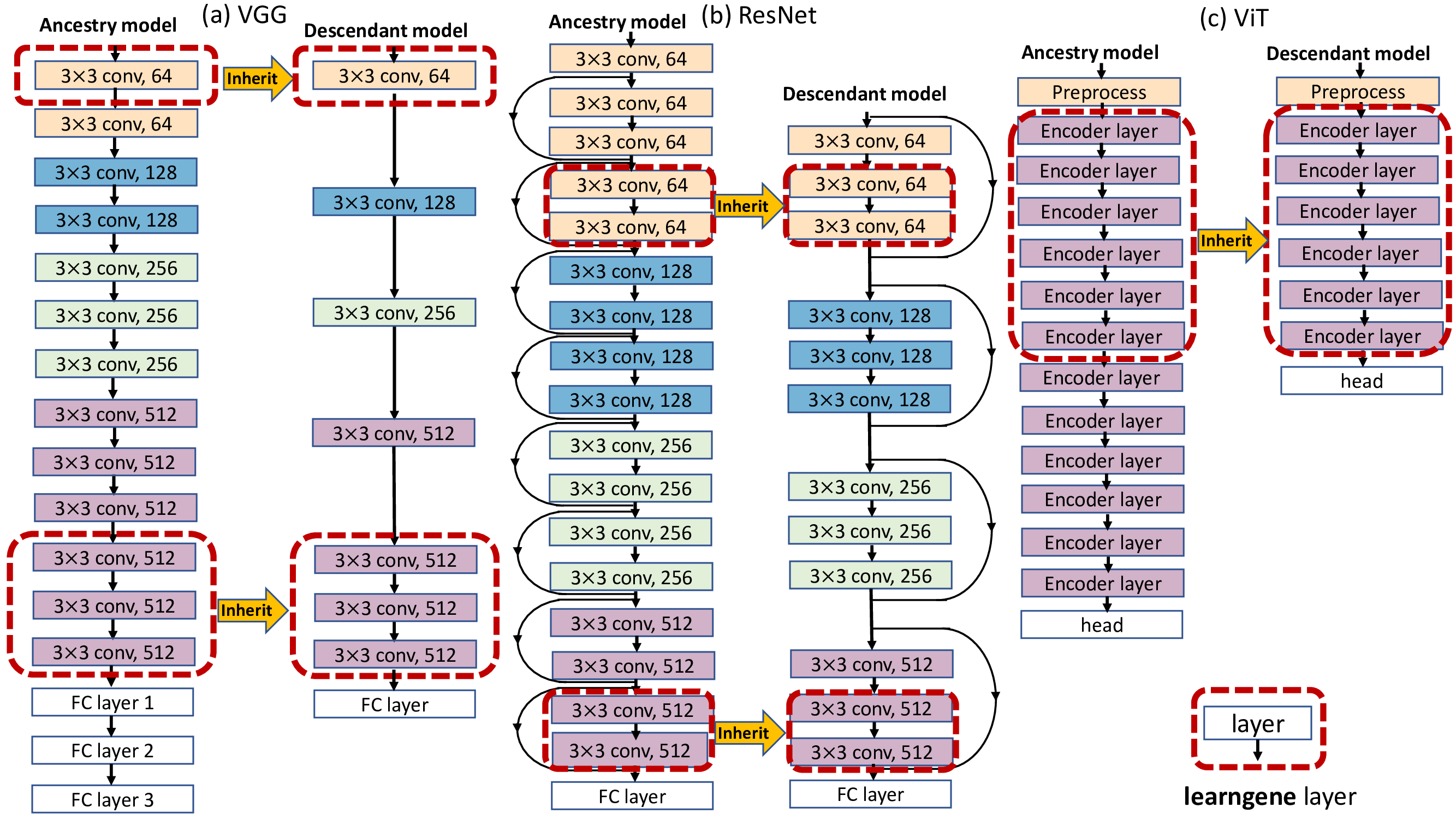}}
\caption{Example network architectures. (a) We pre-train VGG16 as the ancestry model and extract the first layer and the last three layers as the learngene layers. Then, the learngene layers are stacked to a few randomly initialized layers. (b) We pre-train ResNet18 as the ancestry model and extract two convolutional layers with 64 output channels and two convolutional layers with 512 output channels as learngene layers. Similarly, we stack the learngene layers to some intermediate layers (i.e., the convolutional layers with 128 and 256 output channels) and the classification head. (c) For ViT, our algorithm inherits the lower six encoder layers as the learngene layers and stacks them to a classification head of randomly initialized parameters. }
\label{fig:model_description} 
\end{center}
\vspace{-0.2in} 
\end{figure*}

As shown in Figure~\ref{fig:model_description}, we pre-train VGG16 as the ancestry model and extract the first layer and the last three layers as the learngene layers. Then, the learngene layers are stacked onto a few randomly initialized layers. Moreover, we pre-train ResNet18 as the ancestry model and extract two convolutional layers with 64 output channels and two convolutional layers with 512 output channels as learngene layers. Similarly, we stack the learngene layers onto some intermediate layers (i.e., the convolutional layers with 128 and 256 output channels) and the classification head. For ViT, our algorithm selects the lower six encoder layers as the learngene layers and stacks them to a classification head of randomly initialized parameters.

\section{Derivation of Theorem 1}
\label{supp:theorem}

Mathematically, the objective function is given by
\begin{equation}
\label{eq:bi-level-simple}
\begin{aligned}
&\underset{\phi}{\min} \mathcal{L}^{\text {meta}}(\alpha \mid \widehat{\mathcal{D}}) = \frac{1}{M} \sum_{i=1}^{M} \mathcal{L}_{i}^{\text {meta}}(\phi \mid  w^*) \\
&\text { s.t. } w^{*}=\underset{w}{\arg \min } \underbrace{\frac{1}{N} \sum_{i=1}^{N}\left(\mathcal{L}(w \mid \phi^*)+\mathcal{R}_{w, \alpha}\right)}_{\mathcal{L}_{\mathcal{D}}(w, \alpha)} ,
\end{aligned} 
\end{equation}

where $\widehat{\mathcal{D}}$ and $\mathcal{D}$ are meta-data and training data, $\mathcal{L}$ is the loss, $\alpha$ is the output of the meta-network and $\mathcal{R}_{w, \alpha}$ is a regularizer.

\textbf{Theorem 1. } \emph{Suppose the loss function $\mathcal{L}^{\text {meta}}$  satisfy Lipschit smooth and  the Hessian $\nabla^2\mathcal{L}^{\text {meta}}$ is $\rho$-Lipschitz continuous (\ie for every $u, v \in \mathbb{R}^{d}$, $\left\|\nabla^{2} \mathcal{L}^{\text {meta}}(u)-\nabla^{2} \mathcal{L}^{\text {meta}}(v)\right\| \leq \rho\|u-v\|$.). Gradient $\nabla \mathcal{L}^{\text {meta}}$ has a bounded variance w.r.t. meta-data or training data (\ie for any $z$, $\mathcal{B}$, $\mathbb{E}\|\nabla \mathcal{L}^{\text{meta}}(z; \mathcal{B})-\nabla \mathcal{L}^{\text{meta}}(z)\|^{2} \leq \sigma^{2}$ for some $\sigma>0$), and so is the Hessian $\nabla^2\mathcal{L}^{\text {meta}}$. The training function $L$ and the meta objective function $\mathcal{L}^{\text{meta}}$ are nonconvex w.r.t. $w$ and $\phi$ with $\rho$-bounded gradients. Let $\beta_w \in (0, \frac{1}{6 L}]$, and we have }
\begin{equation}
\label{eq:theorem-2}
\frac{1}{T} \sum_{t=0}^{T-1} \mathbb{E} \left[\left\|\nabla \mathcal{L}_{t}^{\text {meta}}(\phi \mid w)\right\|^{2}\right] \leq \mathcal{O}(\frac{L_{\phi}}{\sqrt{T}}),
\end{equation}

where $L_{\phi}$ is some constant independent of the convergence process.

\noindent  \emph{Proof.} 
Since the mini-batch is sampled uniformly from the meta-data, we can rewrite the update equationp as:

\begin{equation}
\label{eq:updatephi-variance}
\begin{aligned}
\phi^{(t+1)}=\phi^{(t)}-\beta_{t}\left[\nabla \mathcal{L}^{\text {meta}}(w^{(t)} \mid \phi^{(t)})+\xi^{(t)}\right] ,
\end{aligned} 
\end{equation}

where $\xi^{(t)}$ are i.i.d random variable with finite variance bounded by $\sigma$ (\ie $\mathbb{E}\left[\left\|\xi^{(t)}\right\|^{2}\right] \leq \sigma^{2}$). For ease of notation, $\beta_{t}$ and $\widehat{\beta_{t}}$ is the learning rate of updating $w$ and $\phi$, where $\beta_{t} = \min \left\{1, \frac{c}{T}\right\}$ for some $c$ and $\widehat{\beta_{t}}=\min \left\{\frac{1}{L}, \frac{\rho}{\sigma \sqrt{T}}\right\}$   

Observe that
\begin{equation}
\label{eq:minus-1}
\begin{aligned}
&\mathcal{L}^{\text {meta }}(w^{(t+1)} \mid \phi^{(t+1)})-\mathcal{L}^{\text {meta }}(w^{(t)}\mid \phi^{(t)}) \\
&=\left\{\mathcal{L}^{\text {meta }}(w^{(t+1)} \mid \phi^{(t+1)})-\mathcal{L}^{\text {meta }}(w^{(t)} \mid \phi^{(t+1)})\right\} \\
&+\left\{\mathcal{L}^{\text {meta }}(w^{(t)} \mid \phi^{(t+1)})-\mathcal{L}^{\text {meta }}(w^{(t)} \mid \phi^{(t)})\right\} .
\end{aligned} 
\end{equation}

Due to the Lipschit smooth of $\mathcal{L}^{\text {meta}}$, we have 

\begin{equation}
\label{eq:minus-2}
\begin{aligned}
&\mathcal{L}^{\text{meta }}(w^{(t+1)} \mid \phi^{(t+1)})-\mathcal{L}^{\text {meta }}(w^{(t)}(\phi^{(t+1)})) \\
&\leq\langle\nabla \mathcal{L}^{\text{meta }}(w^{(t)} \mid \phi^{(t+1)}), (w^{(t+1)} \mid \phi^{(t+1)})-(w^{(t)} \mid \phi^{(t+1)})\rangle \\
&+\frac{K}{2}\|(w^{(t+1)} \mid \phi^{(t+1)})-(w^{(t)} \mid \phi^{(t+1)})\|^{2}.
\end{aligned} 
\end{equation}

Since $\left\|\mathcal{L}^{\text{meta }}(w \mid \phi)\right\| \leq \rho,\left\|L(w \mid \phi)\right\| \leq \rho$, we obtain

\begin{equation}
\label{eq:minus-3}
\begin{aligned}
(w^{(t+1)} \mid \phi^{(t+1)})-(w^{(t)} \mid \phi^{(t+1)}) \leq \beta_t \rho.
\end{aligned} 
\end{equation}

We substitute the Eq.~\eqref{eq:minus-3} into the inequality Eq.~\eqref{eq:minus-2} and then derive:
\begin{equation}
\label{eq:minus-4}
\begin{aligned}
&\|\mathcal{L}^{\text{meta }}(w^{(t+1)} \mid \phi^{(t+1)})-\mathcal{L}^{\text {meta }}(w^{(t)} \mid \phi^{(t+1)})\| \\
&\leq \beta_{t} \rho^{2}+\frac{K \beta_{t}^{2}}{2} \rho^{2}.
\end{aligned} 
\end{equation}

By Lipschitz smoothness of meta loss function, the following holds:

\begin{equation}
\label{eq:minus-5}
\begin{aligned}
&\mathcal{L}^{\text{meta }}(w^{(t)} \mid \phi^{(t+1)})-\mathcal{L}^{\text {meta }}(w^{(t)} \mid \phi^{(t)}) \\
&\leq\langle\nabla \mathcal{L}^{\text{meta }}(w^{(t)} \mid \phi^{(t)}), (w^{(t)} \mid \phi^{(t+1)})-(w^{(t)} \mid \phi^{(t)})\rangle \\
&+\frac{K}{2}\|(w^{(t)} \mid \phi^{(t+1)})-(w^{(t)} \mid \phi^{(t)})\|^{2} \\
&=\langle\nabla \mathcal{L}^{\text{meta }}(w^{(t)} \mid \phi^{(t)}), -\widehat{\beta_{t}}[\nabla \mathcal{L}^{\text{meta }}(w^{(t)} \mid \phi^{(t)})+\xi^{(t)}]\rangle\\
&+\frac{K \widehat{\beta_{t}}^{2}}{2}\|\nabla \mathcal{L}^{\text{meta }}(w^{(t)} \mid \phi^{(t)})+\xi^{(t)}\|_{2}^{2} \\
&=-(\widehat{\beta_{t}}-\frac{K \widehat{\beta_{t}}^{2}}{2})\|\nabla \mathcal{L}^{\text{meta }}(w^{(t)} \mid \phi^{(t)})\|^{2}\\
&+\frac{K \widehat{\beta_{t}}^{2}}{2}\|\xi^{(t)}\|^{2}-(\widehat{\beta_{t}}-K \widehat{\beta_{t}}^{2})\langle \nabla \mathcal{L}^{\text{meta }}(w^{(t)} \mid \phi^{(t)}), \xi^{(t)}\rangle.
\end{aligned} 
\end{equation}

We substitute the Eq.~\eqref{eq:minus-4} and Eq.~\eqref{eq:minus-5} into the inequality Eq.~\eqref{eq:minus-1}and then yield:
\begin{equation}
\label{eq:minus-6}
\begin{aligned}
&\mathcal{L}^{\text {meta }}(w^{(t+1)} \mid \phi^{(t+1)})-\mathcal{L}^{\text {meta }}(w^{(t)}\mid \phi^{(t)}) \\
&\leq \beta_{t} \rho^{2}(1+\frac{\beta_{t} K}{2})-(\widehat{\beta_{t}}-\frac{K \widehat{\beta_{t}}^{2}}{2}) \|\nabla \mathcal{L}^{\text{meta }}(w^{(t)} \mid \phi^{(t)})\|^{2}\\ 
&+\frac{K \widehat{\beta_{t}}^{2}}{2}\|\xi^{(t)}\|^{2}-(\widehat{\beta_{t}}-K \widehat{\beta_{t}}^{2})\langle \nabla \mathcal{L}^{\text{meta }}(w^{(t)} \mid \phi^{(t)}), \xi^{(t)}\rangle.
\end{aligned} 
\end{equation}

Summing up the above inequalities and rearranging the terms, we can obtain
\begin{footnotesize}
\begin{equation}
\label{eq:minus-7}
\begin{aligned}
&\sum_{t=1}^{T}(\widehat{\beta_{t}}-\frac{K \widehat{\beta_{t}}^{2}}{2})\|\nabla \mathcal{L}^{\text{meta }}(w^{(t)} \mid \phi^{(t)})\|^{2}\\ 
&\leq \mathcal{L}^{\text {meta }}(w^{(1)} \mid \phi^{(1)})-\mathcal{L}^{\text {meta }}(w^{(T+1)} \mid \phi^{(T+1)}) +\sum_{t=1}^{T} \beta_{t} \rho^{2}(1+\frac{\beta_{t} K}{2})\\
&- \sum_{t=1}^{T}(\widehat{\beta_{t}}-K \widehat{\beta_{t}}^{2})\langle \nabla \mathcal{L}^{\text{meta }}(w^{(t)} \mid \phi^{(t)}), \xi^{(t)}\rangle + \frac{K}{2} \sum_{t=1}^{T} \widehat{\beta_{t}}^{2}\|\xi^{(t)}\|^{2} \\
&\leq \mathcal{L}^{\text {meta }}(w^{(1)} \mid \phi^{(1)}) + \sum_{t=1}^{T} \beta_{t} \rho^{2}(1+\frac{\beta_{t} K}{2}) \\
&- \sum_{t=1}^{T}(\widehat{\beta_{t}}-K \widehat{\beta_{t}}^{2})\langle \nabla \mathcal{L}^{\text{meta }}(w^{(t)} \mid \phi^{(t)}), \xi^{(t)}\rangle + \frac{K}{2} \sum_{t=1}^{T}\widehat{\beta_{t}}^{2}\|\xi^{(t)}\|^{2}.
\end{aligned} 
\end{equation}
\end{footnotesize}

Since $\mathbb{E}_{\xi}\langle \nabla \mathcal{L}^{\text{meta }}(w^{(t)} \mid \phi^{(t)}), \xi^{(t)}\rangle=0$ and $\mathbb{E}\left[\left\|\xi^{(t)}\right\|^{2}\right] \leq \sigma^{2}$, we take expectations w.r.t. $\xi$ on Eq.~\eqref{eq:minus-7}and then deduce that:

\begin{equation}
\label{eq:minus-8}
\begin{aligned}
&\sum_{t=1}^{T}(\widehat{\beta_{t}}-\frac{K \widehat{\beta_{t}}^{2}}{2})\mathbb{E}_{\xi}\|\nabla \mathcal{L}^{\text{meta }}(w^{(t)} \mid \phi^{(t)})\|^{2} \\
&\leq \mathcal{L}^{\text {meta }}(w^{(1)} \mid \phi^{(1)}) + \sum_{t=1}^{T} \beta_{t} \rho^{2}(1+\frac{\beta_{t} K}{2}) + \frac{K \sigma^{2}}{2} \sum_{t=1}^{T} \widehat{\beta_{t}}^{2}
\end{aligned} 
\end{equation}

Finally,
\begin{tiny}
\begin{equation}
\label{eq:theorem-final}
\begin{aligned}
&\frac{1}{T} \sum_{t=0}^{T-1} \mathbb{E} \left[\left\|\nabla \mathcal{L}_{t}^{\text {meta}}(\phi \mid w)\right\|^{2}\right] \\
&\leq \frac{\sum_{t=1}^{T}\left(\widehat{\beta_{t}}-\frac{K \widehat{\beta_{t}}^{2}}{2}\right) \mathbb{E}_{\xi}\left\|\nabla \mathcal{L}_{t}^{\text {meta}}(\phi \mid w)\right\|^{2}}{\sum_{t=1}^{T}\left(\widehat{\beta_{t}}-\frac{K \widehat{\beta_{t}}^{2}}{2}\right)}\\
&\leq \frac{1}{\sum_{t=1}^{T}\left(2 \widehat{\beta_{t}}-K \widehat{\beta_{t}}^{2}\right)}\left[2\mathcal{L}^{\text {meta }}(w^{(1)} \mid \phi^{(1)}) + \sum_{t=1}^{T} \beta_{t} \rho^{2}(2+\beta_{t}K) + K \sigma^{2} \sum_{t=1}^{T} \widehat{\beta_{t}}^{2}\right]\\
&\leq \frac{1}{\sum_{t=1}^{T} \widehat{\beta_{t}} }\left[2\mathcal{L}^{\text {meta }}(w^{(1)} \mid \phi^{(1)}) + \sum_{t=1}^{T} \beta_{t} \rho^{2}(2+\beta_{t}K) + K \sigma^{2} \sum_{t=1}^{T} \widehat{\beta_{t}}^{2}\right]\\
&\leq \frac{1}{T \widehat{\beta_{t}} }\left[2\mathcal{L}^{\text {meta }}(w^{(1)} \mid \phi^{(1)}) + c \rho^{2}(2+cK) + K \sigma^{2} \sum_{t=1}^{T} \widehat{\beta_{t}}^{2}\right]\\
&\leq \frac{\left[2\mathcal{L}^{\text {meta }}(w^{(1)} \mid \phi^{(1)}) + c \rho^{2}(2+cK)\right]}{T}\frac{1}{\widehat{\beta_{t}} } + K \sigma^{2} \widehat{\beta_{t}}\\
&= \frac{\left[2\mathcal{L}^{\text {meta }}(w^{(1)} \mid \phi^{(1)}) + c \rho^{2}(2+cK)\right]}{T}\max \left\{L, \frac{\sigma \sqrt{T}}{\rho}\right\} + K \sigma^{2} \min \left\{\frac{1}{L}, \frac{\rho}{\sigma \sqrt{T}}\right\}\\
&\leq \frac{\sigma\left[2\mathcal{L}^{\text {meta }}(w^{(1)} \mid \phi^{(1)}) + c \rho^{2}(2+cK)\right]}{\rho \sqrt{T}} + \frac{K \sigma \rho}{\sqrt{T}}  =\mathcal{O}\left(\frac{L_{\phi}}{\sqrt{T}}\right) .
\end{aligned} 
\end{equation}
\end{tiny}
The proof is completed.

\ifCLASSOPTIONcompsoc
  \section*{Acknowledgments}
\else
  \section*{Acknowledgment}
\fi

We sincerely thank Tiankai Hang and Jun Shu for the helpful discussion.

\ifCLASSOPTIONcaptionsoff
  \newpage
\fi



\bibliographystyle{IEEEtran}
\bibliography{IEEEabrv,bare_jrnl_compsoc}
%



%





\begin{IEEEbiography}[{
\includegraphics[width=1in,height=1.25in,clip,keepaspectratio]{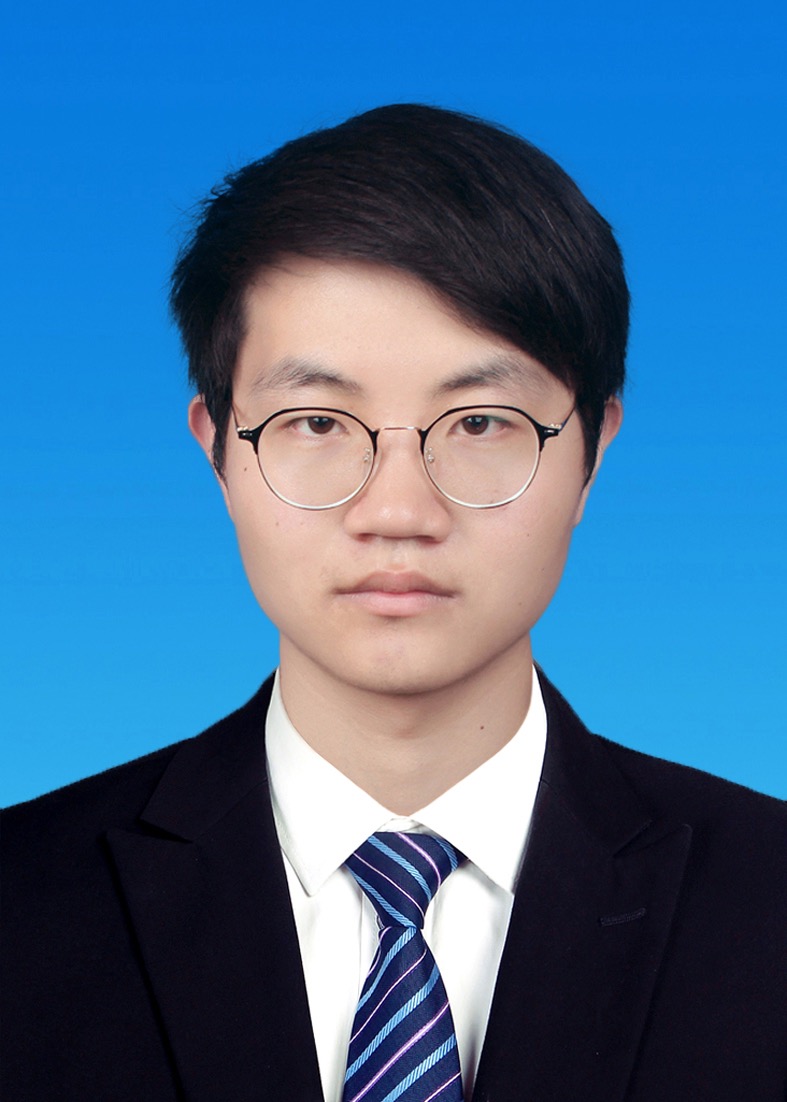}}]{Qiufeng Wang}
 received the B.E. from Southwest Jiaotong University (SWJTU), China, in 2019. He is currently a Ph.D student of Department of School of Computer Science and Engineering, Southeast University, China. His research interests include meta-learning, transfer learning and AutoML.
\end{IEEEbiography}

\begin{IEEEbiography}[{
\includegraphics[width=1in,height=1.25in,clip,keepaspectratio]{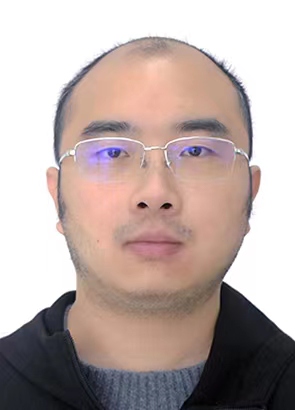}}]{Xu Yang}
 received the B.Eng. degree in Communication Engineering from Nanjing University of Posts and Telecommunications in 2013, the M.Eng. degree in Information Processing from Southeast University in 2016, and the Ph.D. degree in computer science from Nanyang Technological University in 2021. He is currently an Associate Professor at the School of Computer Science and Engineering of Southeast University, China. His research interests mainly include computer vision and machine learning.
\end{IEEEbiography}

\begin{IEEEbiography}[{
\includegraphics[width=1in,height=1.25in,clip,keepaspectratio]{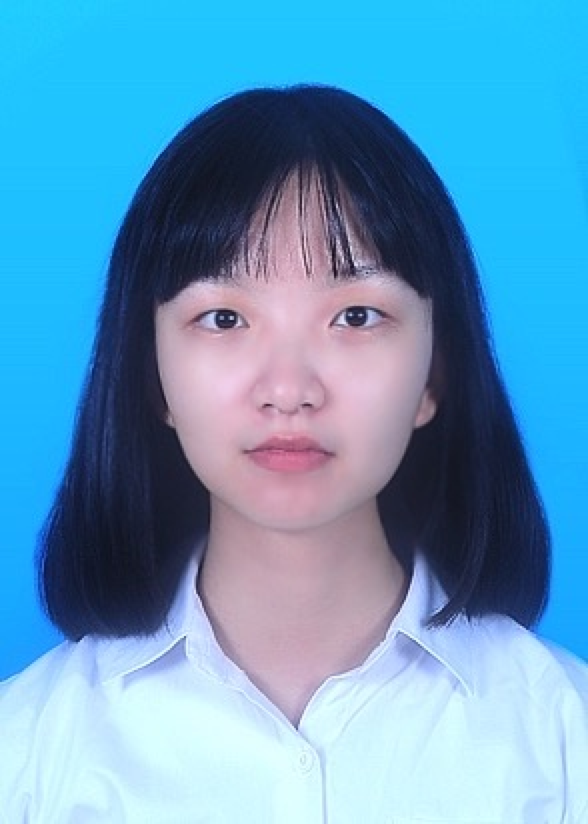}}]{Shuxia Lin}
 received the B.Eng. degree in Computer Science from Southeast University, China, in 2020. She is currently working toward the PhD degree from the School of Computer Science and Engineering of Southeast University, China. Her research interests mainly include computer vision and machine learning.
\end{IEEEbiography}


\begin{IEEEbiography}[{
\includegraphics[width=1in,height=1.25in,clip,keepaspectratio]{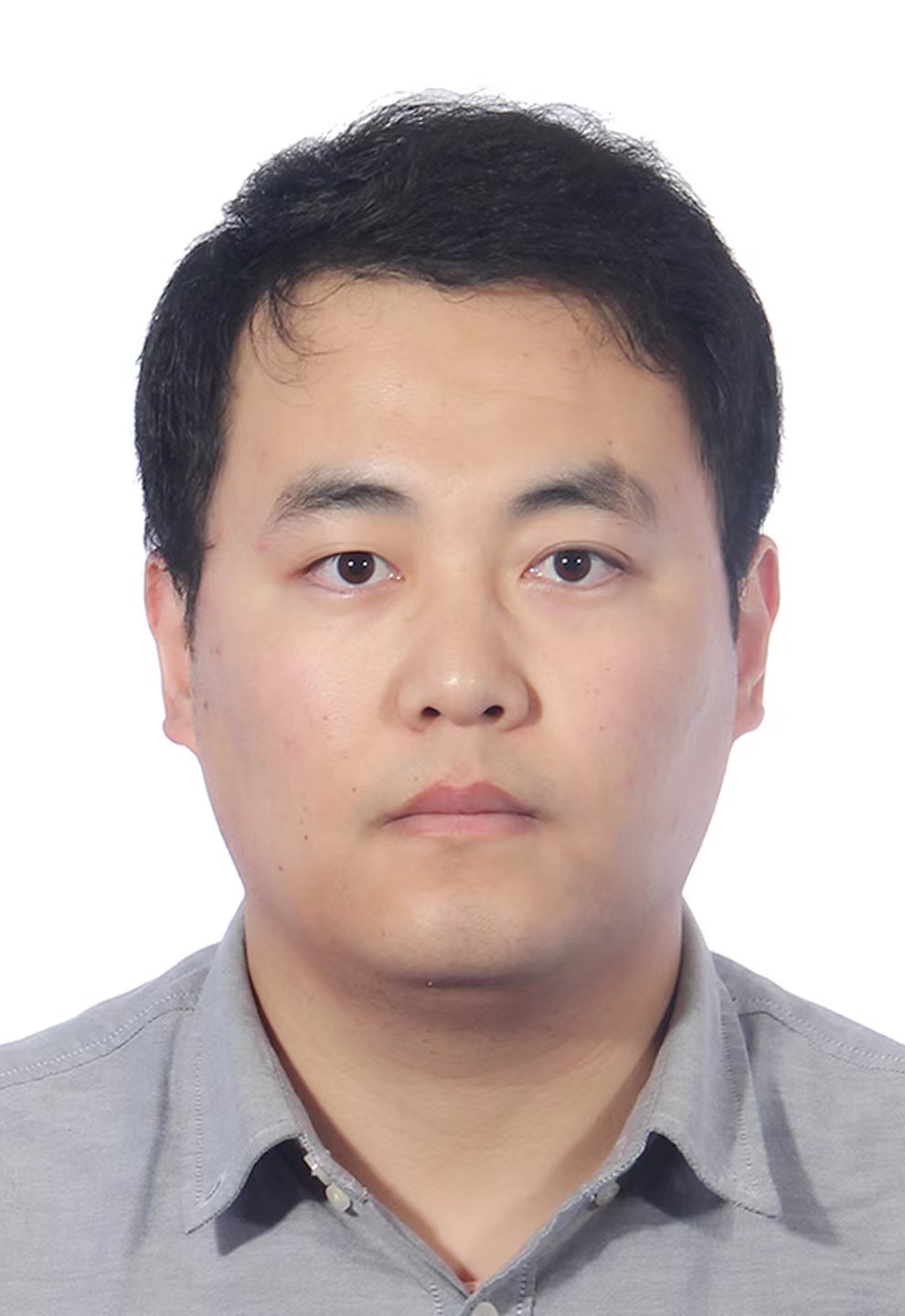}}]{Jing Wang}
   received the B.Sc. degree in computer science from Suzhou University of Science and Technology, Suzhou, China, in 2013, and the M.Sc. degree in computer science from Northeastern University, Shenyang, China, in 2015, and the Ph.D. degree in software engineering from Southeast University, Nanjing, China, in 2021. He is currently an assistant professor  of the School of Computer Science and Engineering, Southeast University, Nanjing. His research interests include pattern recognition and machine learning.
\end{IEEEbiography}


\begin{IEEEbiography}[{
\includegraphics[width=1in,height=1.25in,clip,keepaspectratio]{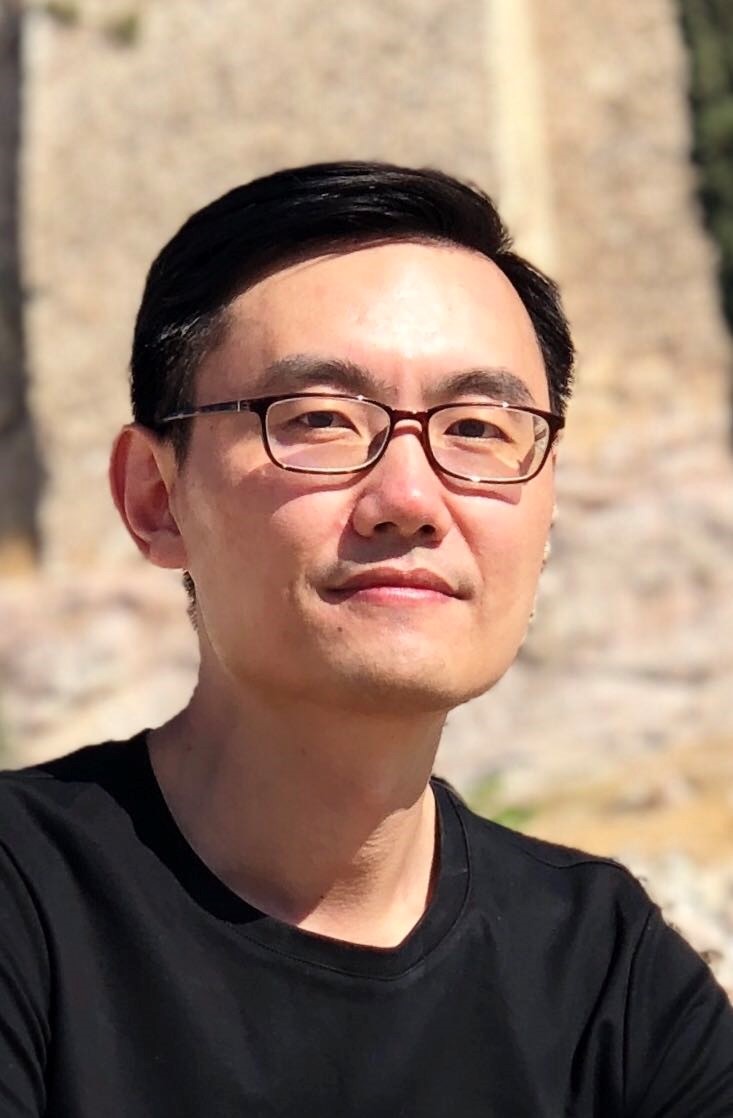}}]{Xin Geng}
     is a chair professor of School of Computer Science and Engineering at Southeast University, China. He received the B.Sc. (2001) and M.Sc. (2004) degrees in computer science from Nanjing University, China, and the Ph.D. (2008) degree in computer science from Deakin University, Australia. His research interests include machine learning, pattern recognition, and computer vision. He has published over 100 refereed papers in these areas, including those published in prestigious journals and top international conferences. He has been an Associate Editor of IEEE T-MM, FCS and MFC, a Steering Committee Member of PRICAI, a Program Committee Chair for conferences such as PRICAI'18, VALSE'13, etc., an Area Chair for conferences such as IJCAI, CVPR, ACMMM, ICPR, and a Senior Program Committee Member for conferences such as IJCAI, AAAI, ECAI, etc. He is a Distinguished Fellow of IETI and a Senior Member of IEEE.
\end{IEEEbiography}




\end{document}